\documentclass[a4paper,fleqn]{cas-dc}

\usepackage{wasysym}
\usepackage[numbers]{natbib}
\let\printorcid\relax
\usepackage{amsmath,amssymb,amsfonts}
\usepackage{algorithmic}
\usepackage{graphicx}
\usepackage{algorithm,algorithmic}
\usepackage{hyperref}
\usepackage{textcomp}
\usepackage{subfigure}
\usepackage{multirow}
\usepackage{bm}
\usepackage{makecell}
\usepackage{adjustbox}

\usepackage{pifont}
\usepackage{hyperref}
\usepackage{color}
\usepackage{url}
\usepackage{booktabs}
\usepackage{soul}
\def\tsc#1{\csdef{#1}{\textsc{\lowercase{#1}}\xspace}}
\tsc{WGM}
\tsc{QE}

\begin{document}
\begin{sloppypar}
\let\WriteBookmarks\relax
\def\floatpagepagefraction{1}
\def\textpagefraction{.001}

\shorttitle{ProCNS: Progressive Prototype Calibration and Noise Suppression for Weakly-Supervised Segmentation}    

\shortauthors{Y. Liu et al.} 

\title [mode = title]{ProCNS: Progressive Prototype Calibration and Noise Suppression for Weakly-Supervised Medical Image Segmentation}  


%
\author[1]{Yixiang Liu}
\fnmark[1]

\author[1,2,3]{Li Lin}
\fnmark[1]

\author[2]{Kenneth K. Y. Wong}
\fnmark[1]

\author[1,3]{Xiaoying Tang}
\cormark[1]

\affiliation[1]{organization={Department of Electronic and Electrical Engineering, Southern University of Science and Technology},
            addressline={1088 Xueyuan Avenue, Nanshan District}, 
            city={Shenzhen},
            postcode={518055}, 
            state={GuangDong},
            country={China}}

\affiliation[2]{organization={Department of Electrical and Electronic Engineering, University of Hong Kong.},
            addressline={1088 Xueyuan Avenue, Nanshan District}, 
            city={Hong Kong},
            country={China}}

\affiliation[3]{organization={Jiaxing Research Institute, Southern University of Science and Technology},
            addressline={Xiuzhou District}, 
            city={Jiaxing},
            postcode={314000}, 
            state={ZheJiang},
            country={China}}

\cortext[1]{Corresponding authors}

\fntext[1]{Co-first authors: Yixiang Liu and Li Lin contributed equally to this work.}


\begin{abstract}
  Weakly-supervised segmentation (WSS) has emerged as a solution to mitigate the conflict between annotation cost and model performance by adopting sparse annotation formats (e.g., point, scribble, block, etc.). Typical approaches attempt to exploit anatomy and topology priors to directly expand sparse annotations into pseudo-labels. However, due to lack of attention to the ambiguous boundaries in medical images and insufficient exploration of sparse supervision, existing approaches tend to generate erroneous and overconfident pseudo proposals in noisy regions, leading to cumulative model error and performance degradation. In this work, we propose a novel WSS approach, named ProCNS, encompassing two synergistic modules devised with the principles of progressive prototype calibration and noise suppression. Specifically, we design a Prototype-based Regional Spatial Affinity (PRSA) loss to maximize the pair-wise affinities between spatial and semantic elements, providing our model of interest with more reliable guidance. The affinities are derived from the input images and the prototype-refined predictions. Meanwhile, we propose an Adaptive Noise Perception and Masking (ANPM) module to obtain more enriched and representative prototype representations, which adaptively identifies and masks noisy regions within the pseudo proposals, reducing potential erroneous interference during prototype computation. Furthermore, we generate specialized soft pseudo-labels for the noisy regions identified by ANPM, providing supplementary supervision. Extensive experiments on six medical image segmentation tasks involving different modalities demonstrate that the proposed framework significantly outperforms representative state-of-the-art methods. Code and data are available at \href{https://github.com/LyxDLiI/ProCNS}{https://github.com/LyxDLiI/ProCNS}.
  \end{abstract}

\maketitle

\section{Introduction}
\label{sec1}
Medical image segmentation is a fundamental task in computer-aided diagnosis, aiming to delineate critical anatomical or pathological regions for subsequent analyses. In recent years, with the rapid advancement of deep learning, a myriad of medical image segmentation methods have been proposed, showcasing remarkable performance. These approaches focus on designing advanced network architectures or incorporating topological priors, typically relying on fully-supervised learning and greatly benefiting from large-scale annotated datasets with high-quality annotations \cite{lin2021bsda,shit2021cldice}. Nonetheless, collecting and annotating large datasets with dense annotations is exceedingly expensive and time-consuming, especially for medical images, as their annotations necessitate expertise and clinical experience.

\begin{figure}[t]
  \centering
  \includegraphics[width=\columnwidth]{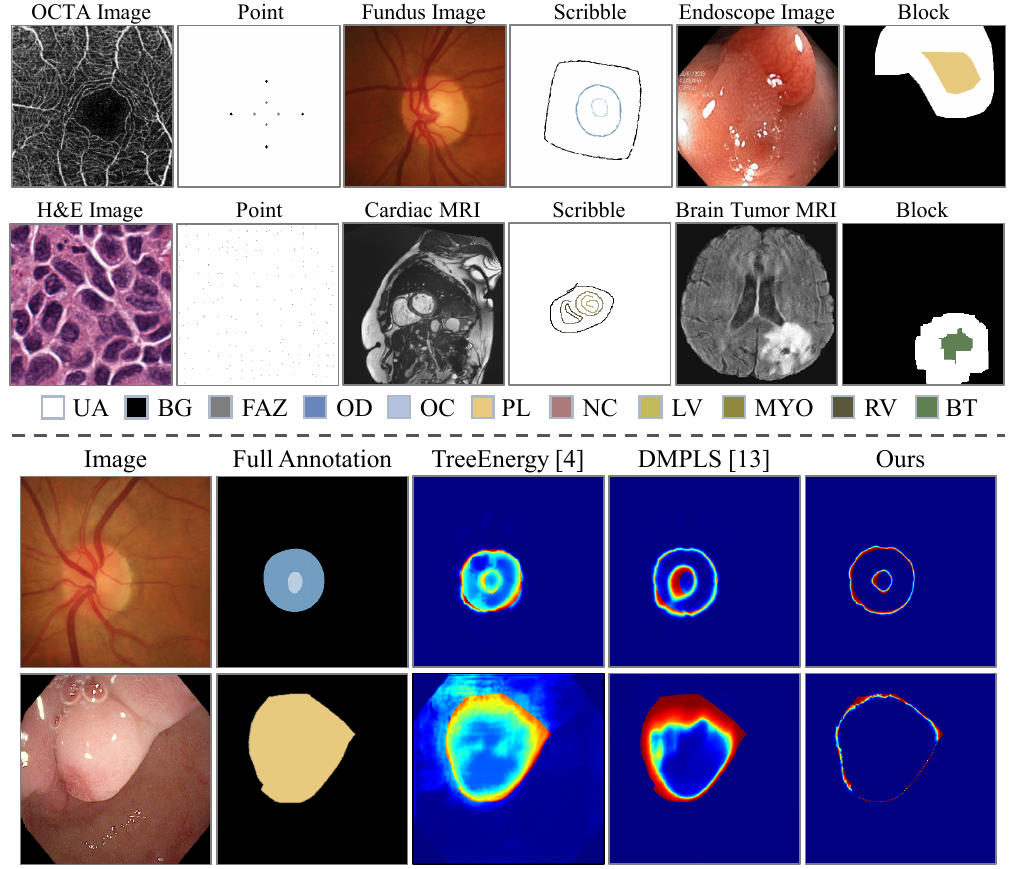}
  \caption{\textbf{Top:} Examples of an optical coherence tomography angiography (OCTA) image, a fundus image, an endoscope image, a hematoxylin and eosin (H\&E)-stained tissue image, a cardiac magnetic resonance image (Cardiac MRI) and a brain tumor magnetic resonance image (Brain Tumor MRI), coupled with their respective sparse annotations of diverse types. UA, BG, FAZ, OD, OC, PL, NC, LV, MYO, RV and BT respectively represent unlabeled region, background, foveal avascular zone, optic disc, optic cup, polyp, nuclei, left ventricle, myocardium, right ventricle and brain tumor. \textbf{Bottom:} Visualization of pseudo-label error maps generated by TreeEnergy \cite{liang2022tree}, DMPLS \cite{luo2022scribble} and our ProCNS.}
  \label{fig:error}
\end{figure}

Weakly-supervised segmentation (WSS) has emerged as a promising solution by employing sparse annotations, such as points, scribbles, blocks and others (as illustrated in the top panel of Fig. \ref{fig:error}), to train segmentation models, effectively alleviating the inherent conflict between annotation cost and model performance. Existing methods can be mainly categorized into pseudo-proposal, consistency learning, auxiliary task and distillation-based methods. Pseudo-proposal methods \cite{wu2023compete, liang2022tree, lin2022yolocurvseg, obukhov2019gated} employ prior knowledge, semantic affinity or model prediction to expand and generate pseudo-labels from the original sparse annotations. They typically involve multi-stage training and are susceptible to noise accumulation. Consistency learning methods \cite{zhang2022cyclemix, ke2021universal} penalize inconsistent predictions on different views of the same image to regularize the training process, yet they fail to exploit the semantic correlation between annotated and unannotated regions. Auxiliary task methods \cite{xu2021leveraging, lee2021weakly} impose comprehensive constraints by incorporating additional tasks such as boundary prediction, which may nevertheless impair the performance of the main segmentation task. Distillation-based methods \cite{chen2021seminar, cheng2023boxteacher} employ teacher models to distill richer knowledge from sparse annotations, transferring it to student models. They inevitably increase model complexity and computational burden.

Among the aforementioned methods, pseudo-proposal approaches are most prevalent. Yet a potentially overlooked pivotal detail is that when sparse labels are generated through either manual annotation or automated algorithms, e.g., Random Walks \cite{grady2006random}, regions selected for annotation tend to be preferentially positioned within readily distinguishable regions (for example, the central regions of the foveal avascular zone and polyp), rather than nebulous and uncertain regions (for example, the boundary intersection regions between the optic disc and the optic cup). Intuitively, those pseudo-labels predominantly inhabit less-informative regions rather than hard-yet-informative ones. The former, easily classified by a model even under the supervision of sparse labels, sharply contrasts the latter; it often exhibits significant prediction fluctuation and unreliability throughout the training process. The skewed annotation proportion, favoring less-informative regions (often the majority) over their hard-yet-informative counterparts (often the minority), may be detrimental to model training. Specifically, under the supervision of such sparse labels, the trained models exhibit a tendency to allocate predictions more extensively to less-informative regions. The diminutive and steady loss values observed in less-informative regions and the pixel-wise averaging characteristic of segmentation losses, e.g., the partial Cross-Entropy (pCE) loss \cite{tang2018normalized}, diminish the efficacy of hard-yet-informative regions, subsequently leading to erroneous predictions at the boundary regions (as illustrated in the bottom panel of Fig. \hspace{-0.05cm}\ref{fig:error}). In medical images, the structures and lesions tend to be inherently more ambiguous than those in natural images, exacerbating the aforementioned issue. Direct or indirect utilization of those erroneous predictions as pseudo-labels may induce further error accumulation, leading to performance degradation.

The most direct solution to the above issue is to increase the coverage and proportion of annotations for hard-yet-informative regions, which, however, conflicts with the objective of reducing manual annotation costs in WSS. Consequently, it is natural to propose prototype representation learning to address the issue. Prototype representation learning has been explored and validated in few-shot and semi-supervised learning tasks \cite{zhu2023transductive,li2021adaptive, yang2020prototype}, which can effectively summarize class representations and generate reliable pseudo-labels. However, in the context of WSS, prototypes extracted from sparse annotations or noisy pseudo-labels lack semantic richness and sufficient accuracy. Inaccurate prototypes may result in misclassifications of unannotated regions. Adaptive approaches that perceive and mask noisy regions while utilizing as many unannotated regions from diverse target classes as possible to generate prototypes could alleviate this issue. However, such explorations are relatively rare.

In such context, we propose a novel weakly-supervised medical image segmentation algorithm, named ProCNS, encompassing two complementary modules conforming to the principles of progressive prototype representation refinement and noise suppression. Firstly, we formulate a Prototype-based Regional Spatial Affinity (PRSA) loss to maximize spatial and semantic pair-wise affinities, thereby providing our model of interest with more robust guidance. The affinities are extracted from the input images and the prototype-refined predictions. Simultaneously, an Adaptive Noise Perception and Masking (ANPM) module is designed to progressively identify and mask noisy regions within the pseudo proposals, mitigating the risk of erroneous interference during prototype computation. In addition, the prototype-refined predictions are harnessed to generate soft pseudo-labels for the noisy regions identified by ANPM, providing additional supervision. The main contributions of this paper are summarized as follows:
        
  \begin{itemize}
      \item To the best of our knowledge, this work is the first attempt to employ progressive prototype calibration and noise suppression to address the insufficiency of prototype semantic representativeness and richness in WSS. Moreover, the proposed ProCNS can be flexibly utilized as a seamless integration plugin for existing WSS methods.
      \item We integrate prototype learning and affinity to propose the PRSA loss, aiming at enhancing the representations' intra-class compactness and inter-class separability by utilizing low-level spatial and high-level semantic pair-wise affinities from the input images and the prototype-refined predictions.
      \item We propose the ANPM module that progressively identifies and masks noisy regions while identifying reliable target regions for prototype calibration. It can also guide the generation of tailored soft pseudo-labels for noisy regions, thus enabling additional supervision.
      \item We evaluate ProCNS on six different medical image segmentation tasks involving various forms of sparse annotations. Experimental results showcase the superiority of ProCNS over existing comparative methods.
  \end{itemize}

\section{Related Works}
\subsection{Weakly-supervised Segmentation}
Weakly-supervised segmentation aims to reduce the annotation cost by training segmentation models on data with inexact annotations. Existing methods fall into four main categories: pseudo-proposal, consistency learning, auxiliary task and distillation-based methods. Pseudo-proposal methods \cite{liang2022tree, zhang2021adaptive, lin2023unifying, ahn2018learning} employ prior knowledge, semantic similarity or model predictions to propagate sparse labels to unlabeled regions, thereby generating extended pseudo-labels. For instance, Liang \emph{et al.} \cite{liang2022tree} employ the minimum spanning tree property to design a tree filter for effectively mitigating pseudo-labels' noise. Compete-to-win \cite{wu2023compete} compares multiple confidence maps produced by auxiliary branches to vote for the best one to serve as the pseudo-label. Consistency learning methods \cite{zhang2022cyclemix, ke2021universal} generally utilize cross-view consistency to penalize inconsistent segmentation. For example, Zhang  \emph{et al.} \cite{zhang2022cyclemix} adopt a mixup strategy to obtain images with diverse views, followed by a consistency loss to regularize the model training process. Auxiliary task methods \cite{Chang_2020_CVPR, xu2021leveraging, yang2023self} enhance comprehensive constraints by integrating other tasks, such as sub-category exploration \cite{Chang_2020_CVPR} and multi-label image classification \cite{xu2021leveraging}. Additionally, Yang \emph{et al.} \cite{yang2023self} propose an innovative self-supervised auxiliary task based on contrastive learning to facilitate the downstream segmentation task. Distillation-based methods \cite{chen2021seminar, zhang2023weakly2, cheng2023boxteacher, fang2023reliable} employ teacher models to distill more richer knowledge from sparse annotations, subsequently transferring to student models. For instance, in the context of WSS, Zhang \emph{et al.} \cite{zhang2023weakly2} develop a self-dual teaching architecture that leverages two-fold information cues, namely the discriminative object region and the full object region, to generate high-quality pseudo-labels, thereby better guiding the training of the student model.

However, due to the lack of specific attention to the ambiguous boundaries in medical images, these methods' performance is generally restricted by the representation bias and the accumulated noise. On the contrary, our approach can alleviate these issues by progressively calibrating prototypes and providing specialized supervision for noisy regions.

\subsection{Prototype Representation Learning}
Given its capacity of clustering similar units into a unified embedding space, prototype representation learning can effectively capture the structures and features of data. It has been comprehensively studied and validated in few-shot, semi-supervised and unsupervised tasks \cite{cermelli2021prototype, feng2023unsupervised, li2023acseg, cheng2023prior, zhou2023advclip, wang2024unlearnable, zhang2024badrobot}. The effectiveness of utilizing prototypes to refine predictions has already been preliminarily explored. For instance, Xu \emph{et al.} \cite{xu2022semi} introduce a multi-prototype classifier to replace the traditional parameterized classifier, while Zhang \emph{et al.} \cite{zhang2023self} utilize sample-wise prototypes to generate cross-sample probability predictions. However, in the WSS setting, the target prototypes generated from sparse labels may lack semantic richness. Directly or indirectly utilizing them for prediction refinement could potentially result in overconfident errors. The idea of adaptively selecting regions with less noise to generate prototypes holds the promise of mitigating this issue. Currently, such explorations are relatively rare and our work aims to fill this research gap.
\vspace{-0.2cm}
\section{Methodology}
\begin{figure*}[t]
\centering
\includegraphics[width=1\textwidth]{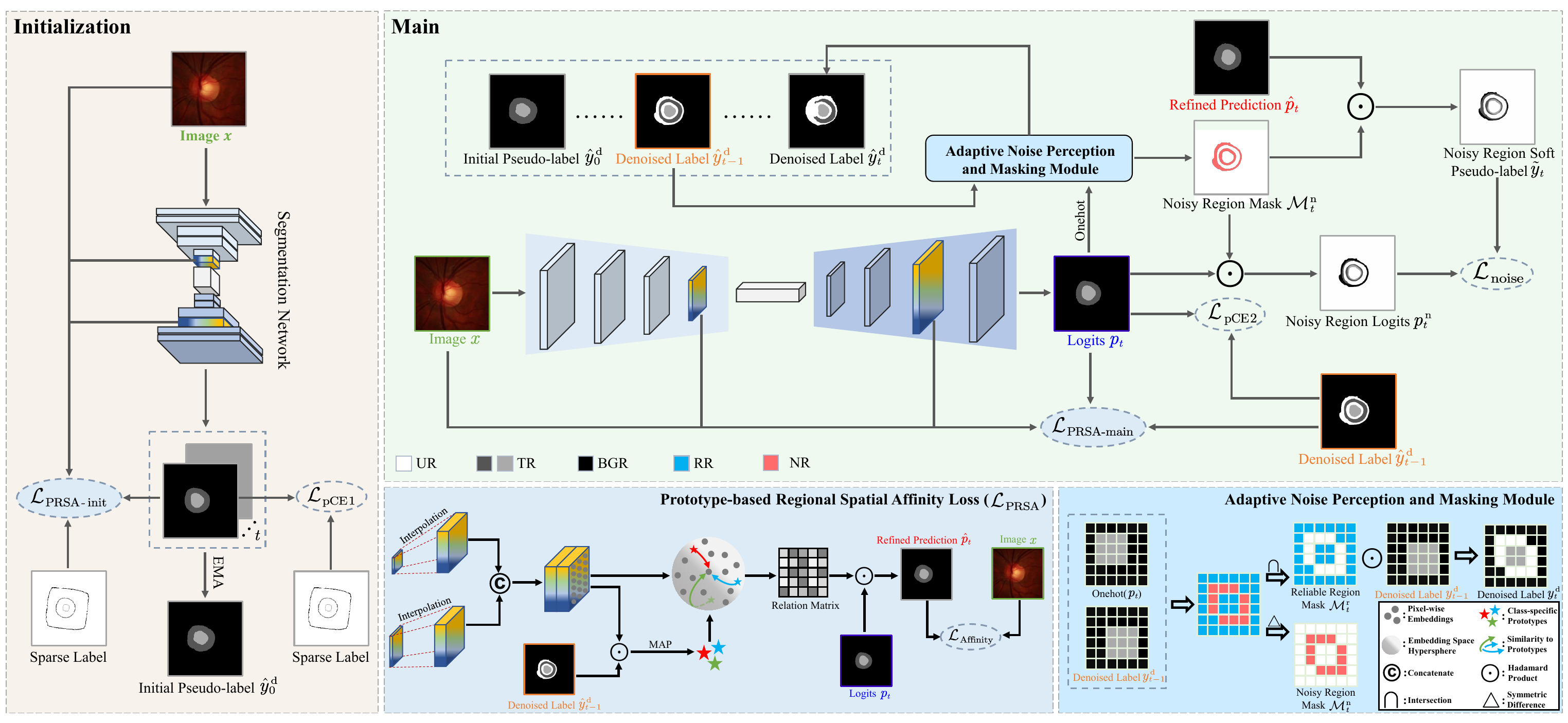} 
\caption{An overview of ProCNS. UR, TR, BGR, RR and NR respectively represent the unlabeled region, target region, background region, reliable region and noisy region. Onehot and MAP respectively denote One-hot-encoding and masked average pooling. In the Initialization stage, a preliminary segmentation model is trained using the sparsely-annotated dataset to generate initial pseudo-labels. In the Main stage, the model is further fine-tuned using dense pseudo-labels. The Main stage consists of two crucial components: the PRSA loss and ANPM.}
\vspace{-0.5cm}
\label{fig:framework}
\end{figure*}

\vspace{-0.1cm}
Our ProCNS framework features itself with a joint training process, as illustrated in Fig. \ref{fig:framework}. The Initialization stage involves utilizing sparse annotations to attain relatively reliable initial pseudo-labels and a preliminary segmentation model. The Main stage comprises two key components: a Prototype-based Regional Spatial Affinity (PRSA) loss and an Adaptive Noise Perception and Masking (ANPM) module. The former utilizes prototypes to optimize pair-wise affinities between images and predictions, while the latter aims to generate more accurate prototypes. In addition, we provide additional soft supervision for noisy regions.

\subsection{Preliminary}
\vspace{-0.1cm}
In the context of WSS, given a sparsely-annotated dataset $\mathcal{D}_{\rm s}=\left\{\left(x_i, y^{\rm s}_i\right) \mid 1 \leq i \leq N\right\}$, where $x_i \in \mathbb{R}^{H \times W}$ is the ${i}$th sample image of size ${H \times W}$ and $y^{\rm s}_i \in\{0,1\}^{H \times W \times c}$ is the corresponding sparse label. In the \textbf{Initialization} stage, $\mathcal{D}_{\rm s}$ is used to train a preliminary segmentation model and deliver initial pseudo-labels. In the \textbf{Main} stage, the training set becomes the corresponding dataset with denoised and dense annotations $\mathcal{D}_{\rm d}=\left\{\left(x_i, \hat{y}^{\rm d}_{t,i}\right) \mid 0 \leq t \leq T, 1 \leq i \leq N\right\}$, where $\hat{y}^{\rm d}_{t, i} \in\{0,1\}^{H \times W \times c}$ denotes the iteratively denoised pseudo-label. Here, $t$ denotes the training epoch at the current stage, $c$ represents the class and $\hat{y}^{\rm d}_{0,i}$ is the $i$th sample's initial pseudo-label obtained in the Initialization stage. The core objective of WSS is to thoroughly exploit the weakly annotated dataset and train a dense segmentation model, maximizing the performance-cost ratio.
\subsection{Generating Initial Pseudo-Labels via Temporal Ensembling}
\vspace{-0.1cm}
As the initial pseudo-labels serve as the benchmark for denoised labels in the Main stage, it is desirable for them to be as reliable as possible. However, due to the imbalanced annotation proportion between less-informative and hard-yet-informative regions, a segmentation model commonly exhibits pronounced predictive uncertainty throughout the entire training process. Consequently, directly employing individual predictions generated from a single model as pseudo-labels is suboptimal. Inspired by Temporal Ensembling \cite{laine2016temporal}, we perform exponential moving average (EMA) of the model's predictions to relieve the issue. The temporal ensemble prediction $\overline{p}_t$ at epoch $t$ is defined as
\begin{equation}
        \overline{p}_t=\alpha p_t+(1-\alpha )\overline{p}_{t-1}, t \in\{1, \ldots, T\},
\end{equation} where $\alpha$ denotes the EMA decay rate and $T$ represents the maximum value of the training epochs in the Initialization stage. The initial pseudo-label $\hat{y}^{\rm d}_{0}$ is generated utilizing $\overline{p}_t$ at $t=T$,
\begin{equation}
      \hat{y}^{\rm d}_{{0}}={\operatorname{argmax}} (\overline{p}_{t=T}).
\end{equation}
\subsection{Prototype-based Regional Spatial Affinity Loss}
As mentioned above, models trained by sparse labels are poorly calibrated and may output overconfident predictions that lack both intra-class compactness and inter-class discrepancy. Inspired by prototype representation learning and the TreeEnergy approach \cite{liang2022tree}, we propose a PRSA loss that leverages low-level spatial and high-level semantic pair-wise affinities to address the mentioned concern.

\subsubsection{Multi-scale Sample-wise Prototype}
Considering that deep embeddings encompass more generalized global semantics while shallow features encapsulate specific local spatial information, the integration of multi-level information often leads to superior performance in medical image segmentation tasks. Moreover, in scenarios such as endoscope images, wherein illumination variation exerts a substantial influence, there may exist substantial distribution gaps among different samples within the same domain or the same dataset. Utilizing class prototypes computed at the dataset or batch level might restrict diversity and compromise the representation capacity. As such, we generate sample-wise prototypes $z_{i,c }$ by integrating multi-scale embeddings, where $i$ and $c$ respectively denote the $i$th sample and the class $c$. Specifically, given an image $x$ and the corresponding label $y$ in a same batch (with a batch size $b$), the deep prototypes $z^{\rm dp}_{c}$ are batch-wise, signifying that images within the same batch share identical deep prototypes. The shallow prototypes $z^{\rm sw}_{i,c }$ are sample-wise, ensuring distinct prototypes for different images. They are calculated via masked average pooling, 
\begin{equation}
\label{eq:pro1}
\begin{aligned}
& z^{\rm dp}_{c}=\frac{1}{\sum_{i=1}^b{\left| \mathcal{W}_{i,c} \right|}}\sum_{i=1}^b{\sum_m^{HW}\mathbb{I}_{\left[ y_{i,m} =c \right]}}f_{i,m}^{\rm h}; \\
& z_{i,c}^{\rm sw}=\frac{1}{\left| \mathcal{W}_{i,c} \right|}\sum_m^{HW}\mathbb{I}_{\left[ y_{i,m}  =c \right]}f_{i,m}^{\rm l}; \\
& z_{i,c}= \operatorname{cat}\left(z^{\rm dp}_{c}, z^{\rm sw}_{i,c }\right),
\end{aligned}
\end{equation}
where $m$ is a pixel and $\operatorname{cat}$ is the concatenation operation through the broadcast mechanism. $f^{\rm h}$ and $f^{\rm l}$ are respectively high-level and low-level embeddings. $\mathbb{I}$ is the indicator function. $\mathcal{W}_{i,c} $ represents the set of pixels belonging to class $c$ in the $i$th sample's label. $f^{\rm h}_i$ and $f^{\rm l}_i$ are interpolated with bilinear interpolation to match the dimension of $y_i$, prior to the calculation.

\subsubsection{Relation Matrix and Prototype-refined Prediction}
A relation matrix $r\in \mathbb{R}^{H\times W\times c}$ is formed by evaluating the degree of correlation between pixel-embeddings and prototype vectors. Given the $i$th sample's embedding $f_i$ generated by concatenating $f^{\rm h}_i$ and $f^{\rm l}_i$, we compute the correlation strength between each prototype $z_{i,c }$ and the pixel-embedding $f_{i,m}$ via the cosine similarity 
\begin{equation}
\operatorname{sim}\left(f_{i,m}, z_{i,c}\right) =\frac{f_{i,m}^{\rm T} \cdot z_{i,c}}{||f_{i,m} ||\cdot ||z_{i,c}||}, 
\end{equation}
where $\rm T$ is the transpose operation. And $r_{i}$ is defined as
\begin{equation}
r_i=\mathcal{N} \left( \operatorname{ReLU}\left(\operatorname{sim}\left(f_i, z_{i,c}\right) \right) \right),
\end{equation}
where $\operatorname{ReLU}$ is the regular ReLU function, i.e., $\operatorname{ReLU}\left( x \right)=x$ if $x > 0$ and $\operatorname{ReLU}\left( x \right) = 0$ otherwise. $\mathcal{N}$ is the 1-dimensional normalization function. We utilize the logits $p_i$ and the relation matrix $r_i$ to form the prototype-refined prediction $\hat{p_i}$,
\begin{equation}
\hat{p}_i = \operatorname{Softmax} \left( p_i\cdot r_i \right),
\end{equation}
where $\operatorname{Softmax}$ is applied over the class channel $c$.

\subsubsection{Affinity Loss}
Our affinity loss is designed to facilitate the propagation of region-level semantic information from reliable regions to noisy ones by maximizing the affinity between pixels. Feeding the image $x$ to the model, the model outputs the logits $p$.

Following Zhang \emph{et al.} \cite{zhang2023zscribbleseg}, we utilize a Gaussian kernel function to design the low-level weight function $\omega^{\rm low}$, which is defined by the distinction between two pixels within the image $x$ in terms of image intensity value $v$ and spatial location $l$,

\begin{equation}
\omega^{\rm low}(m,n)=\exp \left\{ -\frac{\left( l_m-l_n \right) ^2}{2\sigma _{l}^{2}}-\frac{\left( v_m-v_n \right) ^2}{2\sigma _{v}^{2}} \right\} ,
\end{equation}
where $n$ represents a pixel at a different location from $m$, while $\sigma _{l}$ and $\sigma _{v}$ are respectively the bandwidth parameter for location and intensity. The high-level weight function $\omega^{\rm high}$ is defined as 
\begin{equation}
\omega^{\rm high}(m)  = \hat{p}_m. 
\end{equation}
The pair-wise affinity matrices denote as $A^{\rm low} $ and $A^{\rm high}$
\begin{equation}
\begin{aligned}
& A_{m}^{\rm low}=\sum_{n\in \mathcal{W}^r_m \backslash\{m\}}w^{\rm low}(m,n) ;\\
& A_{m}^{\rm high}=\sum_{n\in \mathcal{W}^r_m \backslash\{m\}}{w^{\rm high}(n)}, 
\end{aligned}
\end{equation}
where $\mathcal{W}^r_m$ represents the set of pixels within the $r \times r$ neighborhood centered around $m$. Here, $r$ represents the radius. The affinity loss is defined as
\begin{equation}
\mathcal{L} _{\mathrm{Affinity}}=-\frac{1}{HW}\sum_m^{HW}\sum_c\left( {\left( \sum_c^{}{A_{m,c}^{\text{high}}} \right)}\cdot A_{m}^{\text{low}}\cdot \hat{p}_{m,c} \right). 
\end{equation}
\subsection{Adaptive Noise Perception and Masking Module}
The initial pseudo-labels could carry noise, especially near boundaries and background regions resembling the foreground. Prototypes from such labels are inaccurate and can mislead model representation. Observations by \cite{liu2022adaptive} indicate that a model's adaptation or fitting to noisy labels is gradual, wherein the model initially fits correct labels, then gradually overfits to noise. Employing initial pseudo-labels to train a model until it reaches a bottleneck or a turning point in performance and then automatically refining the labels may facilitate the model in transcending these limitations. 

As such, we employ an iterative label refinement strategy to design ANPM, which adaptively extracts the reliable region mask $\mathcal{M}^{\rm r}$ and the noisy region mask $\mathcal{M}^{\rm n}$ of a denoised label $\hat{y}^{\rm d}_{t-1}$ and the corresponding prediction $p_{t}$. Subsequently, the reliable region is preserved for prototype computation, while the noisy region is masked out, as illustrated in the lower panel of Fig. \ref{fig:framework}. The calculation goes as follows
\begin{equation}
\begin{aligned}
\mathcal{M}^{\rm r}_{t}&={\operatorname{Onehot}} (p_{t}) \cap \hat{y}^{\rm d}_{t-1}; \\ \mathcal{M}^{\rm n}_{t}&={\operatorname{Onehot}} (p_{t})\triangle \hat{y}^{\rm d}_{t-1},
\end{aligned}
\end{equation}
where $\operatorname{Onehot}$ is the One-hot-encoding operation, $\cap$ and $\triangle$ respectively denote intersection and symmetric difference.

We utilize $\mathcal{M}^{\rm r}_t$ to generated the denoised label $\hat{y}^{\rm d}_{t}$ at epoch $t$, 
\begin{equation}
  \hat{y}^{\rm d}_{t}=\mathcal{M}^{\rm r}_{t}\odot \hat{y}^{\rm d}_{t-1}. 
\end{equation}
We then iteratively calibrate the prototypes by substituting $\hat{y}^{\rm d}_{t}$ into Eq. (\ref{eq:pro1}).
\subsection{Masked Region Reassignment and Soft Supervision}
We additionally craft tailored soft pseudo-labels for regions marked as noise by ANPM, offering additional supervision. Regarding the masked noisy region $\mathcal{M}^{\rm n}_{t}$ at epoch $t$, we utilize the relation matrix derived from the calibrated prototypes to reassign pixels at the corresponding regions in the original prediction. Specifically, we employ $\mathcal{M}^{\rm n}_{t}$ to extract a prediction $p^{\rm n}_{t}$ and the corresponding soft pseudo-label $\tilde{y}_{t}$ targeted at the noisy region, which are expressed as
\begin{equation}
\begin{aligned}
    p^{\rm n}_{t} &= \mathcal{M}^{\rm n}_{t}\odot p_{t}; \\  \tilde{y}_{t} &= \mathcal{M}^{\rm n}_{t}\odot \hat{p}_{t}.
\end{aligned}
\end{equation}
Subsequently, the soft label is utilized to supervise the corresponding region in the network's initial prediction. Here, we adopt the Dice loss, denoted as $\mathcal{L}_{\rm noise}$, to achieve this supervision, 
\begin{equation}
    \mathcal{L}_{\rm noise}=1-\frac{2|p^{\rm n}_{t}\cap \tilde{y}_{t}|}{|p^{\rm n}_{t}|+|\tilde{y}_{t}|}.
\end{equation}
\subsection{Total Objective Formulation}\label{subsec:IntegrablePlugins}
Following Lee \emph{et al.} \cite{lee2020scribble2label} and Zhang \emph{et al.} \cite{zhang2022cyclemix}, we employ the pCE loss \cite{tang2018normalized} to provide direct supervision by matching predictions with sparse labels, which is expressed as 
\begin{equation}
\mathcal{L}_{\rm pCE }=-\frac{1}{\left|\mathcal{W}_{\rm L}\right|} \sum_{\forall m \in \mathcal{W}_{\rm L}} y_m^{\rm s} \log \left(p_m\right), 
\end{equation}
where $m$ represents a pixel, $\mathcal{W}_L$ is the set of labeled pixels in the sparse label and $|\cdot|$ denotes the number of pixels.

Our ProCNS can be divided into two stages. In the first stage, we utilize $\mathcal{L}_{\rm pCE_1 }$ and $\mathcal{L}_{\rm PRSA-init}$ to train the preliminary model, wherein $\mathcal{L}_{\rm PRSA-init}$ relies solely on the prototypes computed from sparse labels $y^{\rm s}$. The overall loss goes as follows
\begin{equation}
    \mathcal{L}_{\rm {init }}=\mathcal{L}_{\rm pCE_1 }+\lambda_1 \mathcal{L}_{\rm PRSA-init},
\end{equation}
where $\lambda_{\rm 1}$ is a trade-off coefficient between the two losses. In ProCNS, the Initialization phase solely aims to acquire relatively reliable initial pseudo-labels from a preliminary model, which can be adapted by any WSS method. In other words, the core components of ProCNS's Main stage can also serve as seamlessly integrable subsequent plugins for other WSS methods. 

During the Main stage, we employ $\mathcal{L}_{\rm PRSA-main}$, which utilizes prototypes computed based on the iteratively updated denoised labels $y^{\rm d}_t$. In addition to the use of $\mathcal{L}_{\rm pCE_1 }$ for sparse labels, we also integrate $\mathcal{L}_{\rm pCE_2 }$ which pertains to the denoised labels. Furthermore, with $\mathcal{L}_{\rm noise}$ providing additional supervision over noisy regions, the total loss can be defined as
\begin{equation}
\label{eq:mainloss}
    \mathcal{L}_{\rm {main }}=\mathcal{L}_{\rm pCE_1 } + \lambda_{\rm 2} \mathcal{L}_{\rm pCE_2 } + \lambda_{\rm 3} \mathcal{L}_{\rm PRSA-main} + \lambda_{\rm 4} \mathcal{L}_{\text {noise }}, 
\end{equation}
where $\lambda_{\rm 2}$, $\lambda_{\rm 3}$, $\lambda_{\rm 4}$ are trade-off coefficients among the different losses.

Notably, the Main stage's model is derived by further training the model from the Initialization stage, wherein the training epochs for the latter approximately account for one-tenth of the total training epochs. Furthermore, ANPM achieves prototype calibration by progressively replacing the labels $y$ used for prototype computation in Eq. \ref{eq:pro1} (i.e., replacing $\hat{y}^{\rm d}_{t-1}$ with $\hat{y}^{\rm d}_{t}$ at epoch $t$), rather than introducing extra training parameters. Collectively, these points indicate that the network architectures and the quantity of the to-be-optimized parameters remain constant throughout the entire training process. Thus, ProCNS can be treated as an end-to-end framework that inter-connects two training paradigms.
\begin{table}[b]
  \caption{Summary of the six sparsely-annotated datasets. The proportion denotes the average percentage of the sparsely-annotated regions over the fully-annotated ones. The relative time cost (RTC) denotes the approximate percentage of the average time consumed in manually annotating images with sparse and full annotations.}
  \label{tab:data_detail}
  \centering
  \renewcommand\arraystretch{1.4}
  \resizebox{\columnwidth}{!}{
  \begin{tabular}{ccccc}
    \hline
    Modality            & Dataset     & Format   & Proportion & RTC        \\ \hline
    OCTA                & sOCTA       & Point    & 1.22\%     & $\sim$9\%  \\
    Fundus              & RIM-ONE     & Scribble & 10.39\%    & $\sim$17\% \\
    Endoscope           & Kvarsir-SEG & Block    & 61.47\%    & $\sim$21\% \\
    H\&E                & WO          & Point    & 0.15\%     & $\sim$2\%  \\
    Cardiac MRI         & ACDC        & Scribble & 11.03\%    & $\sim$19\% \\
    Brain Tumor MRI     & BraTS2019   & Block    & 64.70\%    & $\sim$24\% \\ \hline
    \end{tabular}
  }
\end{table}
\section{Experiments and Results}
\subsection{Datasets and Evaluation Metrics}
We evaluate our approach on the \textbf{sOCTA} \cite{wang2021deep}, \textbf{RIM-ONE} \cite{fumero2011rim}, \textbf{Kvarsir-SEG} \cite{jha2020kvasir}, \textbf{WO} \cite{kumar2017dataset}, \textbf{ACDC} \cite{bernard2018deep} and \textbf{BraTS2019} \cite{bakas2018identifying} datasets, utilizing point, scribble and block annotations for the FAZ, ODOC, polyp, nuclei, cardiac multi-structures (CM) and whole brain tumor (WT) segmentation tasks (as delineated in Table \ref{tab:data_detail}). Among these six tasks, the cardiac multi-structures segmentation task employs manual scribble annotations provided by Valvano \emph{et al.} \cite{valvano2021learning}, while all others employ sparse annotations generated using different automated algorithms based on their original full annotations.

The \textbf{sOCTA}, \textbf{RIM-ONE}, \textbf{Kvarsir-SEG} and \textbf{WO} datasets consist of 708, 99, 900 and 16 2D training samples, as well as 304, 60, 100 and 8 2D testing samples. The \textbf{ACDC} dataset consists of 80 3D training samples and 20 3D testing samples. For these five datasets, we randomly split 20\% of the training samples for validation. The \textbf{BraTS2019} dataset consists of 335 3D samples, each containing four modalities: FLAIR, T1, T1ce and T2. Following Luo \emph{et al.} \cite{luo2022semi}, we perform weakly-supervised whole brain tumor segmentation using only FLAIR images. The \textbf{BraTS2019} samples are randomly split into 250 for training, 25 for validation and 60 for testing.

For data pre-processing, the OCTA images, fundus images and endoscope images are respectively resized to $256 \times 256$, $384 \times 384$ and $384 \times 384$ as inputs. Following Yao \emph{et al.} \cite{yao2024position} and Qu \emph{et al.} \cite{qu2019weakly}, the H\&E-stained tissue images are first segmented into $250 \times 250$ patches from the original size of $1000 \times 1000$. The patches are then resized to $1024 \times 1024$ using bicubic interpolation to serve as inputs. Following previous works \cite{luo2022scribble} \cite{agravat2019brain} \cite{lin2024fedlppa}, we convert the 3D volumes of cardiac MRI and brain tumor MRI into 2D slices (slices not containing the corresponding target are excluded) and the 2D slices are respectively resized to $256 \times 256$ and $192 \times 192$ as inputs. Then, we normalize the intensity values of all images to have zero mean and unit variance. Additionally, random rotation and flipping are applied for data augmentation \cite{lin2024fedlppa}.

For evaluation, we employ the commonly-used Dice coefficient (DSC) and 95\% Hausdorff distance (HD95) to quantitatively evaluate the segmentation performance.

\subsection{Implementation Details}
All compared WSS methods and ProCNS are implemented with PyTorch on Nvidia RTX 3090 GPUs. We employ the vanilla UNet \cite{ronneberger2015u} as the network architecture, with its encoder and decoder respectively comprising four downsampling blocks and four upsampling blocks. The highest-dimensional embedding is 256. The SGD optimizer with a momentum of $0.9$ and a weight decay of $1e^{-4}$ is adopted. The polynomial decay schedule is employed. The initial learning rate for the FAZ, ODOC, polyp tasks is uniformly set to $1 \times 10^{-2}$, while for the nuclei, CM and WT tasks, it is respectively set to $5 \times 10^{-2}$, $5 \times 10^{-2}$ and $1 \times 10^{-1}$. The EMA decay rate is set to $0.8$. The trade-off coefficients $\lambda_1$, $\lambda_2$, $\lambda_3$ and $\lambda_4$ are respectively set to $0.1, 0.5, 0.1$ and $0.01$. The $\sigma _{l}$, $\sigma _{v}$ and radius $r$ of the neighborhood window are respectively set to 6, 0.1 and 5. The number of the training epochs during the Initialization stage is respectively set to 100, 400, 100, 5, 40 and 25 for the FAZ, ODOC, polyp, nuclei, CM and WT segmentation tasks.


We now provide detailed discourse regarding the acquisition of the three forms of sparse annotations in the following text. The generation of sparse annotations is facilitated by three automated algorithms, which are implemented utilizing OpenCV2, SimpleITK and Scikit-image. For points, inspired by Kim \emph{et al.} \cite{kim2023devil}, Zhang \emph{et al.} \cite{zhang2023weakly} and Yao \emph{et al.} \cite{yao2024position}, we automatically generate points from full annotations. Specifically, we extract the maximum inscribed rectangles from the objects. The four sides of each rectangle are contracted inward by a predetermined value. For the FAZ task, we extract the midpoints of the four sides, whereas for the nuclei task, we extract the rectangle's center. These points, encompassing multiple pixels, are then convolved with a discrete 2D Gaussian kernel to simulate a manual brushstroke. For scribbles, following Luo \emph{et al.} \cite{luo2022scribble}, a cross-shaped kernel is utilized to perform morphological erosion from the center pixel of each full annotation (through the \textit{erode} function in OpenCV2). Subsequently, the residual regions of disparate classes are compressed into thin lines (through the \textit{skeletonize} function in Scikit-image), thus generating centerlines or skeletons as scribble annotations. Following Liang \emph{et al.} \cite{liang2022tree}, block annotations are derived via morphological erosion transformations starting from the edges and progressing towards the center.

To assess the sparse annotation cost, we report the proportion of the sparsely-annotated region over the fully-annotated one and the relative time cost provided by clinical physicians in Table \ref{tab:data_detail}. Notably, although the proportion of the block annotation in the endoscope dataset exceeds 60\%, employing automatic internal filling makes their actual time cost essentially equivalent to that of the scribble annotation.

\subsection{Ablation Studies and Analyses}
\subsubsection{Ablation Studies of ProCNS}\label{subsec:ab_procns}

\begin{table}[t]
\caption{Ablation study on the three key components in ProCNS regarding DSC. Here, $\checkmark$ indicates the corresponding component is applied. “$*$” represents $\rm p \leq 0.05$ in a Wilcoxon matched-pairs signed rank test comparing the DSC of the pertinent component before and after application. The column for ODOC reports the average DSC of OD and OC. The best results are in bold.}
\centering
\label{tab:ablation1}
\resizebox{\columnwidth}{!}{
\begin{tabular}{ccc|ccc}
\hline
$\boldsymbol{\mathcal{L}}_{\rm \textbf{PRSA}}$ & \textbf{ANPM} & $\boldsymbol{\mathcal{L}}_{\rm \textbf{noise}}$ & \textbf{FAZ}   & \textbf{ODOC}  & \textbf{Polyp} \\ \hline
-    & -    & -     & 90.02 & 87.39 & 78.99 \\ \hline
$\checkmark$  &      &       & 92.38$^{*}$\hspace{-0.15cm} & 87.91 & 79.46$^{*}$\hspace{-0.15cm} \\
$\checkmark$  & $\checkmark$  &       & 93.30$^{*}$\hspace{-0.15cm}  & 88.05 & 80.55 \\
$\checkmark$  & $\checkmark$  & $\checkmark$   & \textbf{93.74}$^{*}$\hspace{-0.15cm} & \multicolumn{1}{r}{\textbf{88.32}$^{*}$}\hspace{-0.15cm} & \textbf{82.73}$^{*}$\hspace{-0.15cm} \\ \hline
\end{tabular}
}
\end{table}
\begin{table}[b]
\caption{Ablation study on the initial PRSA loss regarding DSC. The column for ODOC reports the average DSC of OD and OC.}
\centering
\label{tab:ablation_add_initialprsaloss}
\renewcommand\arraystretch{1.5}
  \begin{tabular}{c|ccc}
    \hline
    Method                       & FAZ   & ODOC  & Polyp \\ \hline
    ProCNS w/o $\boldsymbol{\mathcal{L}}_{\rm \textbf{PRSA-init}}$ & 92.60 & 87.62 & 80.51 \\
    ProCNS                       & 93.74 & 88.32 & 82.21 \\ \hline
    \end{tabular}
\end{table}

To ascertain the efficacy of the three key components $\mathcal{L}_{\rm PRSA}$, ANPM and $\mathcal{L}_{\rm noise}$ within ProCNS, we perform a series of ablation studies. In this context, the baseline is defined as the segmentation model that is trained with sparse annotations and solely employs the pCE loss.

Then, we incrementally incorporate the three components into the baseline. Table \ref{tab:ablation1} presents the quantitative results from the ablation analyses on the three tasks in terms of DSC. Compared to the baseline, $\mathcal{L}_{\rm PRSA}$ respectively improves model performance by 2.36\%, 0.52\% and 0.47\% for the FAZ, ODOC and polyp segmentation tasks. The integration of ANPM yields DSC improvements of 0.92\%, 0.14\% and 1.09\%. Lastly, the addition of $\mathcal{L}_{\rm noise}$ results in further DSC enhancements by 0.44\%, 0.29\% and 2.18\%. These results demonstrate the effectiveness of all components, particularly $\mathcal{L}_{\rm noise}$, which significantly enhances the performance across all three segmentation tasks ($p \leq 0.05$).

\subsubsection{Ablation Studies of Initial PRSA Loss}
To ascertain the efficacy of utilizing the initial PRSA loss at the Initialization stage, we report the results on the FAZ, ODOC and polyp tasks with and without the initial PRSA loss, as shown in Table \ref{tab:ablation_add_initialprsaloss}. All results with the initial PRSA loss outperform those without it, indicating that the initial PRSA loss is beneficial for model performance. This may be attributed to the effective utilization of the high-level semantic correlation between unlabeled and sparsely-annotated regions, which aids the preliminary model in capturing boundary features and achieves a more reliable preliminary model for the Main stage, consequently assisting the final model in more accurately segmenting targets.

\begin{figure}[b]
    \footnotesize
      \subfigure{\includegraphics[width=0.33\linewidth]{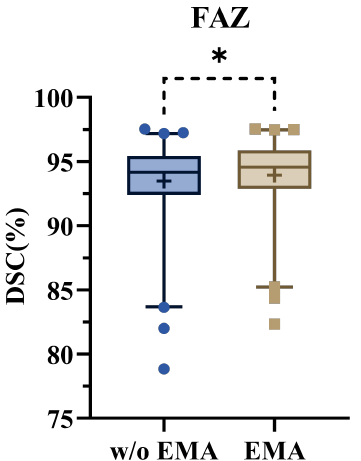}}\hspace{-0.18cm}
      \subfigure{\includegraphics[width=0.33\linewidth]{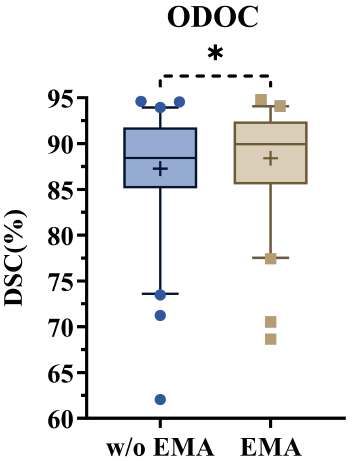}}\hspace{-0cm}
      \subfigure{\includegraphics[width=0.33\linewidth]
      {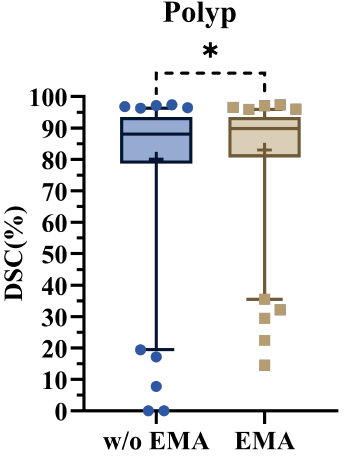}}\hspace{-0.08cm}
      \\
      
    \caption{Ablation analysis results of the temporal ensembling strategy at the Initialization stage with regards to DSC. “+” is the average DSC, the central line indicates the median DSC and data points $\Circle$  $\Square$ are outliers. “$*$” indicates $\rm p \leq 0.05$ from a Wilcoxon matched-pairs signed rank test.}
    \vspace{-0.2cm}
    \label{fig:EMA}
  \end{figure}
\begin{figure}[t]
      \centering
      \includegraphics[width=1\columnwidth]{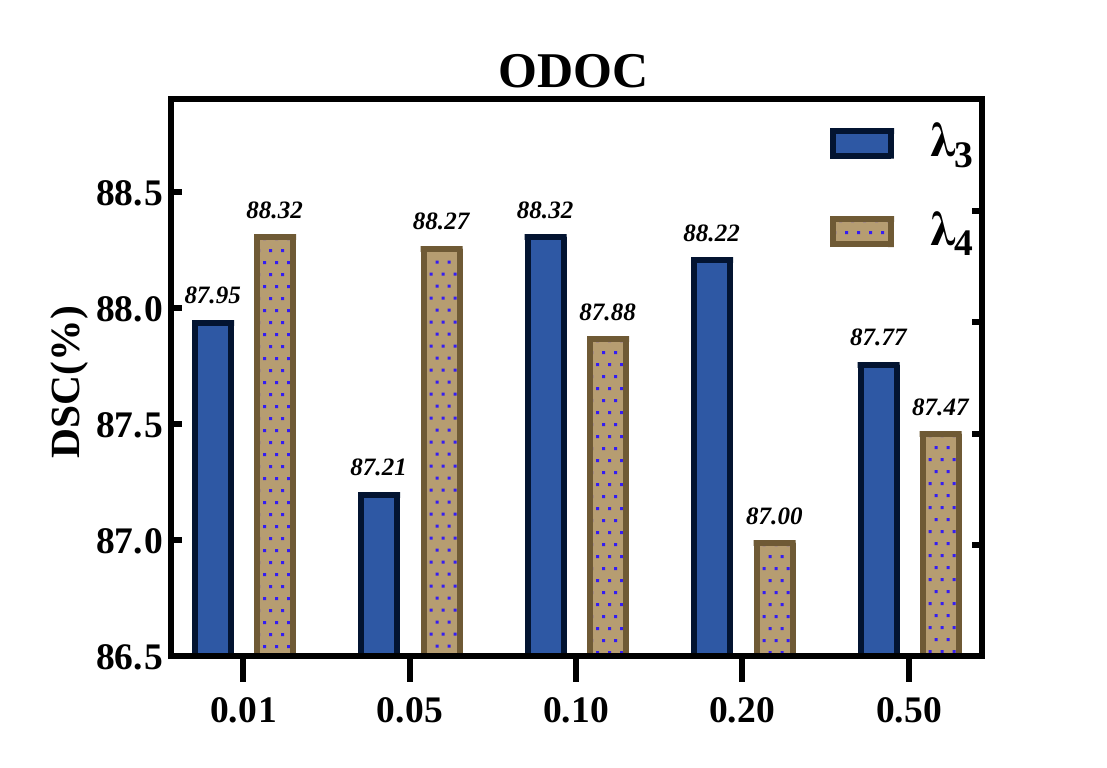}
      \caption{Performance with varied trade-off coefficients $\lambda_3$ and $\lambda_4$.}
      \label{fig:ablation_coefficients}
\end{figure} 
\subsubsection{Ablation Studies of Temporal Ensembling}
We assess the efficacy of utilizing the temporal ensembling strategy (EMA) to generate the initial pseudo-labels during the Initialization stage. As depicted in Fig. \ref{fig:EMA}, employing EMA to update the initial pseudo-labels significantly enhances the model's segmentation performance ($p \leq 0.05$ for all three tasks). These results clearly indicate the effectiveness of the temporal ensembling strategy. This can be attributed to the temporal ensembling's ability to provide tolerance for erroneous predictions in ambiguous regions, which facilitates the generation of more reliable initial pseudo-labels for the Main stage, consequently resulting in more accurate initial prototypes.
\subsubsection{Hyper-parameter and Prototype-refined Strategy Analysis}
\begin{figure}[b]
      \centering
      \includegraphics[width=0.9\columnwidth]{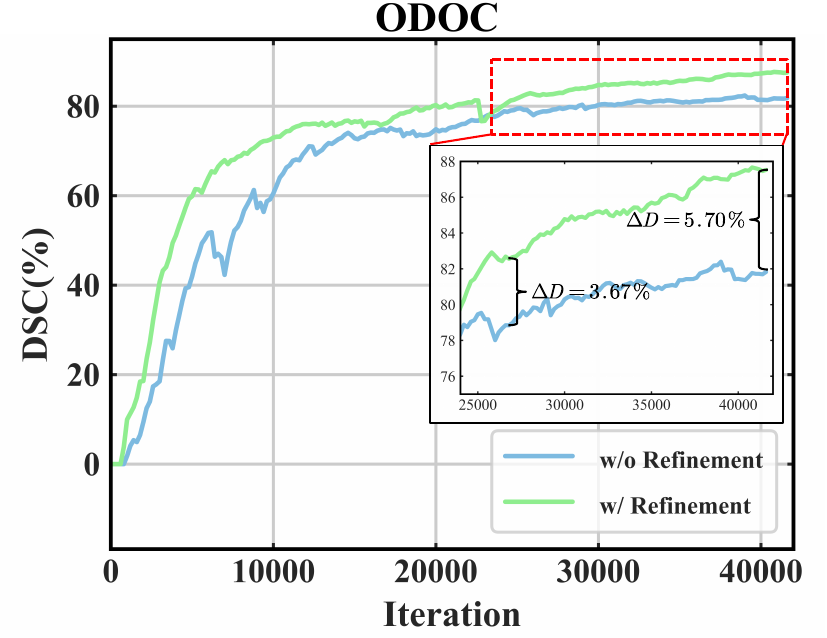}
      \caption{Qualitative evaluation on the prototype-refined predictions. “w/ Refinement” and “w/o Refinement” respectively refer to the outputs of models employing ProCNS with and without using the prototype-refined strategy. $\Delta D$ denotes the DSC difference between “w/o Refinement” and “w/ Refinement”.}
      \label{fig:ablation_refinement}
  \vspace{-0.2cm}
\end{figure}
We evaluate the influence of certain hyper-parameters and the efficacy of utilizing prototype-refined predictions on the ODOC task. The hyper-parameters of interest are the trade-off coefficients $\lambda_3$ and $\lambda_4$ in Eq. (\ref{eq:mainloss}), which are employed to balance $\mathcal{L}_{\rm PRSA-main }$ and $\mathcal{L}_{\text {noise }}$. As shown in Fig. \ref{fig:ablation_coefficients}, $\lambda_3$ is suitable to be set to 0.1 and 0.2 and $\lambda_4$ is suitable to be set to 0.01 and 0.05. The best performance is achieved when $\lambda_3 = 0.1$ and $\lambda_4 = 0.01$. We apply this optimal configuration to the other two tasks as well. The efficacy of utilizing prototype-refined predictions can be indirectly assessed by comparing “w/o Refinement” with “w/ Refinement” in Fig. \ref{fig:ablation_refinement}. With the increase in Iteration, the performance in both settings consistently improves and $\triangle D$ gradually increases from 3.67\% to 5.70\%. These results indicate that extra knowledge can be learned by the segmentation model through the application of prototype-refined predictions.
\begin{table}[t]
  \caption{Ablation analysis results on prototype granularity and scale on the ODOC task. The “en-$i$th” and “de-$i$th” respectively denote the embedding from the $i$th downsampling and upsampling blocks of UNet. The best results are in bold.}
  \label{tab:ablation3_prototype}
  \centering
  \renewcommand\arraystretch{1}
  \tabcolsep=0.3cm{%
  \begin{tabular}{cc|c|cc}
  \hline
  \multicolumn{2}{c|}{\textbf{Prototype}}                    & \textbf{Variant}         & \textbf{DSC}$\uparrow$   & \textbf{HD95}$\downarrow$  \\ \hline
  \multicolumn{2}{c|}{\multirow{3}{*}{Granularity}}       & Batch-wise      & 88.17 & 10.17 \\ \cline{3-5} 
  \multicolumn{2}{c|}{}   & Epoch-wise      & 87.96 & 10.75 \\ \cline{3-5} 
  \multicolumn{2}{c|}{}  & Sample-wise      & \textbf{88.32} & \textbf{9.83} \\  \hline
  \multirow{8}{*}{Scale} & \multirow{4}{*}{Single} & en-4th          & 87.94 & 10.29 \\ \cline{3-5} 
                          &                         & en-3rd          & 87.83 & 10.86 \\ \cline{3-5} 
                          &                         & de-4th          & 87.63 & 10.79  \\ \cline{3-5} 
                          &                         & de-3rd          & 87.56 & 10.42 \\ \cline{2-5} 
                          & \multirow{4}{*}{Multiple}                   & en-4th + de-4th & 88.06 & 9.94  \\ \cline{3-5} 
                          &                         & en-4th + de-3rd & 88.16 & 10.04 \\ 
                          &                         & en-3rd + de-4th & 87.99 & 10.18 \\ \cline{3-5} 
                          &                         & en-3rd + de-3rd & \textbf{88.32} & \textbf{9.83} \\ \hline
  \end{tabular}
  }
\end{table}

\subsubsection{Ablation Studies of Prototype}
\begin{table}[b]
  \caption{Comparison with state-of-the-art WSS methods on computational burden.}
  \label{tab:comparsion2_time}
  \centering
  \renewcommand\arraystretch{1.2}
  \tabcolsep=0.3cm{%
  \begin{tabular}{l|cc}
    \hline
    Method        & \begin{tabular}[c]{@{}c@{}}Training Time\\ (s/epoch)\end{tabular} & \begin{tabular}[c]{@{}c@{}}Testing Effiency\\ (samples/s)\end{tabular} \\ \hline
    $\text{UNet}_{pCE}$          & 4.71                                                              & 39                                                                  \\
    EM \cite{grandvalet2004semi}            & 4.67                                                              & 39                                                                  \\
    Random Walks \cite{grady2006random}  & 2.62                                                              & 38                                                                  \\
    AC \cite{chen2019learning}            & 2.71                                                              & 39                                                                  \\
    GatedCRF \cite{obukhov2019gated}      & 4.86                                                              & 35                                                                  \\
    S2L \cite{lee2020scribble2label}           & 2.98                                                              & 38                                                                  \\
    SeL \cite{chen2021seminar}           & 6.24                                                              & 38                                                                  \\
    USTM \cite{liu2022weakly}          & 16.34                                                             & 38                                                                  \\
    DMPLS \cite{luo2022scribble}         & 4.22                                                              & 33                                                                  \\
    TreeEnergy \cite{liang2022tree}    & 10.80                                                             & 36                                                                  \\
    $\text{S}^2$ME \cite{wang2023s}          & 7.72                                                              & 38                                                                  \\
    ScribbleVC \cite{li2023scribblevc}    & 28.62                                                             & 12                                                                  \\
    ProCNS (Ours) & 10.75                                                             & 38                                                                  \\ \hline
    \end{tabular}
  \vspace{-0.2cm}
  }
\end{table}
We conduct further research into the impact of prototype granularity and prototype scale to substantiate the rationality of employing multi-scale sample-wise prototypes in the design of the PRSA loss. As shown in Table \ref{tab:ablation3_prototype}, for granularity analysis, we compare the ultimate performance of the model using sample-wise, batch-wise and epoch-wise prototypes. The computation methods for the latter two are detailed in \cite{zhang2021prototypical}. Among them, the sample-wise approach achieves the highest DSC of 88.32\%, demonstrating that sample-wise prototypes can adaptively represent class information conforming to the distribution of each sample, thereby effectively mitigating the issue of decreased model generalizability due to data heterogeneity. Regarding prototype scale analysis, we report the training results of single-scale prototypes respectively generated from the embeddings of the 3rd and 4th downsampling as well as the 3rd and 4th upsampling blocks. Furthermore, we showcase the performance of multi-scale prototypes obtained from various combinations of the aforementioned four blocks. All multi-scale combinations outperform their single-scale counterparts, indicating that integrating multi-level information enriches prototype semantics and leads to better performance. The combination of the 3rd downsampling and the 3rd upsampling blocks achieves the highest DSC.

\begin{table*}[h]
  \caption{Comparison with state-of-the-art WSS methods on the FAZ, Polyp and ODOC segmentation tasks. $\text{UNet}_{pCE}$ represents the UNet model trained with sparse annotations, employing only the pCE loss. $\text{UNet}_{F}$ represents the model trained with full annotations, utilizing only the CE loss. “$*$” indicates $\rm p \leq 0.05$ in a Wilcoxon matched-pairs signed rank test when comparing ProCNS with the best-performing method in other WSS methods. The best results are in bold and the second-best results are underlined.}
  \label{tab:comparison}
  \centering
  \renewcommand\arraystretch{1.2}
  \resizebox{\textwidth}{!}{
  \begin{tabular}{l|cc|cc|cccccc}
  \hline
  \multirow{3}{*}{\textbf{Method}}                          & \multicolumn{2}{c|}{\textbf{sOCTA}} & \multicolumn{2}{c|}{\textbf{Kvarsir-SEG}} & \multicolumn{6}{c}{\textbf{RIM-ONE}}                                             \\ \cline{2-11} 
                                                      & \multicolumn{2}{c|}{\textbf{FAZ}}   & \multicolumn{2}{c|}{\textbf{Polyp}}       & \multicolumn{2}{c}{\textbf{OD}} & \multicolumn{2}{c}{\textbf{OC}} & \multicolumn{2}{c}{\textbf{Avg.}} \\ \cline{2-11} 
                                                      & \textbf{DSC}$\uparrow$         & \textbf{HD95}$\downarrow$        & \textbf{DSC}$\uparrow$            & \textbf{HD95}$\downarrow$           & \textbf{DSC}$\uparrow$        & \textbf{HD95}$\downarrow$      & \textbf{DSC}$\uparrow $       & \textbf{HD95}$\downarrow$      & \textbf{DSC}$\uparrow$         & \textbf{HD95}$\downarrow$       \\ \hline
  $\text{UNet}_{pCE}$                                      & 90.02$\pm$5.48       & 29.02$\pm$37.56       & 78.99$\pm$19.97          & 53.02$\pm$59.18          & 93.23$\pm$3.93      & 13.72$\pm$23.83     & 81.55$\pm$11.18      & 14.32$\pm$17.23     & 87.39       & 14.02      \\ \hline
  EM \cite{grandvalet2004semi}     & 92.49$\pm$4.58       & 8.34$\pm$8.48        & \underline{79.48$\pm$19.34}          & 53.90$\pm$57.09          & \underline{94.04$\pm$2.61}      & 10.28$\pm$11.90     & 80.89$\pm$12.01      & 12.54$\pm$7.46     & 87.46       & 11.41      \\
  Random Walks \cite{grady2006random}        & 90.93$\pm$5.12       & 8.35$\pm$4.04        & 73.90$\pm$19.77           & 60.66$\pm$61.53          & 89.90$\pm$4.34       & 13.69$\pm$6.23     & 80.68$\pm$12.79      & 12.75$\pm$8.42     & 85.29       & 13.22      \\
  AC \cite{chen2019learning}        & 92.01$\pm$5.04       & 15.18$\pm$23.13       & 78.49$\pm$20.55          & 54.29$\pm$59.94          & 93.04$\pm$3.92      & 24.37$\pm$41.03     & 80.09$\pm$12.45      & 17.28$\pm$27.94     & 86.56       & 20.82      \\
  GatedCRF \cite{obukhov2019gated}  & 92.39$\pm$4.78       & \underline{7.49$\pm$4.11}        & 79.24$\pm$23.02          & 52.35$\pm$63.03          & 93.12$\pm$2.29      & 9.25$\pm$4.80      & \underline{82.32$\pm$12.05}      & \underline{11.24$\pm$7.38}     & 87.72       & \underline{10.24}      \\
  S2L \cite{lee2020scribble2label} & 91.49$\pm$5.09       & 12.05$\pm$17.97       & 77.17$\pm$19.85          & 55.69$\pm$55.78          & 92.74$\pm$3.81      & 28.30$\pm$44.36      & 81.14$\pm$11.89      & 16.64$\pm$19.16     & 86.94       & 22.47      \\
  SeL \cite{chen2021seminar} & 90.29$\pm$5.47       & 9.61$\pm$5.29       & 76.23$\pm$24.99          & \underline{49.22$\pm$53.86}         & 93.86$\pm$2.74      & \underline{8.37$\pm$3.67}     & 81.27$\pm$12.45      & 12.59$\pm$7.89     & 87.57       & 10.48      \\
  USTM \cite{liu2022weakly}        & 91.20$\pm$4.82       & 9.68$\pm$11.47        & 78.96$\pm$18.41          & 51.03$\pm$54.79          & 93.78$\pm$2.85           & 14.77$\pm$27.54          & 81.30$\pm$12.27           & 15.02$\pm$18.30          & 87.54            & 14.89           \\
  DMPLS \cite{luo2022scribble}    & 92.45$\pm$4.74       & 12.30$\pm$23.02        & 76.93$\pm$22.62          & 55.09$\pm$66.30          & 92.67$\pm$2.87      & 33.59$\pm$66.39     & 80.82$\pm$13.13      & 40.96$\pm$79.34     & 86.74       & 37.27      \\
  TreeEnergy \cite{liang2022tree}   & \underline{92.84$\pm$4.55}       & 7.66$\pm$6.08        & 76.94$\pm$22.75          & 55.61$\pm$62.34          & 93.96$\pm$3.09      & 9.40$\pm$6.20       & 81.31$\pm$12.12      & 12.73$\pm$7.85     & 87.63       & 11.06      \\
  $\text{S}^2$ME \cite{wang2023s}                                                & 92.33$\pm$4.73             & 12.44$\pm$22.80             & 77.26$\pm$20.11                & 57.91$\pm$51.69               & 93.64$\pm$3.52           & 8.80$\pm$4.56         & 82.00$\pm$11.24           & 11.68$\pm$6.32          & 87.82            & \underline{10.24}           \\
  ScribbleVC \cite{li2023scribblevc}                                              & 91.89$\pm$5.14            &  8.21$\pm$4.63           & 79.14$\pm$24.48               & 51.66$\pm$67.33              & 93.58$\pm$3.61           &  10.50$\pm$21.83         &  \textbf{82.60$\pm$14.05}          & \textbf{10.96$\pm$5.73}          &  \underline{88.09}           &  10.73          \\
  ProCNS (Ours)                                       & \textbf{93.74$\pm$4.40}$^{*}$     & \textbf{6.72$\pm$5.83}$^{*}$        & \textbf{82.73$\pm$17.93}$^{*}$           & \textbf{47.80$\pm$53.70}$^{*}$          & \textbf{94.55$\pm$2.84}$^{*}$      & \textbf{7.72$\pm$4.16}$^{*}$      & 82.09$\pm$12.62      & 11.94$\pm$8.29     & \textbf{88.32}       & \textbf{9.83}       \\ \hline
  $\text{UNet}_{F}$                                   & 95.14$\pm$5.46       & 4.80$\pm$3.08         & 
  86.01$\pm$18.40          & 48.72$\pm$52.30          & 95.39$\pm$4.06      & 7.21$\pm$7.39      & 81.70$\pm$12.86       & 12.02$\pm8.21$     & 88.54       & 9.61       \\ \hline
  
  \end{tabular}
}
\vspace{-0.3cm}
\end{table*}
\begin{table*}[h]
  \caption{Comparison with state-of-the-art WSS methods on the WT, Nuclei and CM segmentation tasks. $\text{UNet}_{pCE}$ represents the UNet model trained with sparse annotations, employing only the pCE loss. $\text{UNet}_{F}$ represents the model trained with full annotations, utilizing only the CE loss. “$*$” indicates $\rm p \leq 0.05$ in a Wilcoxon matched-pairs signed rank test when comparing ProCNS with the best-performing method in other WSS methods. The best results are in bold and the second-best results are underlined.}
  \label{tab:comparison_add}
  \centering
  \renewcommand\arraystretch{1.2}
  \resizebox{\textwidth}{!}{
    \begin{tabular}{l|cc|cc|cccccccc}
      \hline
      \multirow{3}{*}{\textbf{Method}}                             & \multicolumn{2}{c|}{\textbf{BraTS2019}}   & \multicolumn{2}{c|}{\textbf{WO}}     & \multicolumn{8}{c}{\textbf{ACDC}}                                                                                       \\ \cline{2-13} 
                                                    & \multicolumn{2}{c|}{\textbf{WT}} & \multicolumn{2}{c|}{\textbf{Nuclei}} & \multicolumn{2}{c}{\textbf{RV}}     & \multicolumn{2}{c}{\textbf{MYO}}    & \multicolumn{2}{c}{\textbf{LV}}    & \multicolumn{2}{c}{\textbf{Avg.}} \\ \cline{2-13} 
                                                    & \textbf{DSC}$\uparrow$             & \textbf{HD95}$\downarrow$           & \textbf{DSC}$\uparrow$          & \textbf{HD95}$\downarrow$         & \textbf{DSC}$\uparrow$         & \textbf{HD95}$\downarrow$         & \textbf{DSC}$\uparrow$         & \textbf{HD95}$\downarrow$         & \textbf{DSC}$\uparrow$         & \textbf{HD95}$\downarrow$        & \textbf{DSC}$\uparrow$        & \textbf{HD95}$\downarrow$        \\ \hline
      $\text{UNet}_{pCE}$                                       & 72.94$\pm$25.43                  & \underline{14.10$\pm$15.35}        & 72.63$\pm$5.42                 & 90.28$\pm$42.37     & 64.13$\pm$26.25 & 131.38$\pm$41.75 & 65.72$\pm$17.51 & 120.80$\pm$36.13 & 79.04$\pm$20.99 & 85.50$\pm$58.87 & 69.63      & 112.56      \\ \hline
      EM \cite{grandvalet2004semi}       & 73.26$\pm$25.52                  & 17.02$\pm$20.95        & 72.30$\pm$8.43                 & \underline{70.25$\pm$40.43}     & 81.72$\pm$26.05 & 14.65$\pm$32.02  & 80.07$\pm$14.95 & 16.76$\pm$31.30  & 87.85$\pm$17.93 & 13.45$\pm$30.53 & 83.22      & 14.95       \\
      Random Walks \cite{grady2006random}          & 78.05$\pm$23.71                  & 19.44$\pm$25.18        & 69.36$\pm$6.56                 & 75.03$\pm$38.79     & 77.50$\pm$27.48 & 7.87$\pm$9.77    & 66.64$\pm$22.88 & 6.81$\pm$7.47    & 78.80$\pm$30.93 & 3.58$\pm$6.98   & 74.31      & 6.09        \\
      AC\cite{chen2019learning}          & 74.33$\pm$23.68                  & 16.19$\pm$17.52        & 73.14$\pm$5.06                 & 91.45$\pm$46.97     & 71.38$\pm$27.11 & 102.47$\pm$52.42 & 68.61$\pm$16.87 & 114.92$\pm$43.36 & 77.87$\pm$23.39 & 63.08$\pm$59.78 & 72.62      & 93.49       \\
      GatedCRF\cite{obukhov2019gated}    & 77.15$\pm$23.45                  & 17.81$\pm$22.91        & \underline{75.34$\pm$4.85}                 & 76.08$\pm$44.35     & 82.05$\pm$25.48 & 7.25$\pm$12.60   & 80.82$\pm$15.01 & \textbf{4.84$\pm$10.59}   & 87.65$\pm$17.71 & 4.54$\pm$9.77   & 83.51      & \textbf{5.54}        \\
      S2L \cite{lee2020scribble2label}   & 70.35$\pm$23.53                  & 17.48$\pm$20.53        & 74.08$\pm$5.43                 & 73.56$\pm$39.23     & 77.67$\pm$25.31 & 59.22$\pm$52.45  & 70.92$\pm$16.72 & 73.55$\pm$50.93  & 78.39$\pm$21.76 & 65.89$\pm$53.53 & 75.66      & 66.22       \\
      SeL \cite{chen2021seminar}   & 76.59$\pm$25.92                 & 14.64$\pm$19.95        & 74.53$\pm$5.64                 & 71.02$\pm$39.51     & 78.84$\pm$29.38 & 6.24$\pm$9.42  & 75.76$\pm$16.50 & 56.37$\pm$58.93  & 85.33$\pm$20.34 & \underline{3.96$\pm$6.73} & 79.98      & 22.19       \\
      USTM \cite{liu2022weakly}          & 71.43$\pm$21.09                  & 15.74$\pm$15.56        & 72.52$\pm$6.12                 & 75.26$\pm$42.46     & 78.52$\pm$26.13 & 56.75$\pm$58.16  & 77.54$\pm$14.74 & 50.15$\pm$52.63  & 84.37$\pm$19.05 & 45.59$\pm$56.97 & 80.15      & 50.83       \\
      DMPLS \cite{luo2022scribble}      & 73.24$\pm$25.11                  & 14.83$\pm$17.71        & 73.26$\pm$6.53                 & 80.48$\pm$41.65     & 81.62$\pm$25.52 & \underline{5.83$\pm$9.39}    & \textbf{82.49$\pm$13.24} & 6.91$\pm$15.27   & \underline{88.67$\pm$17.41} & 4.04$\pm$8.92   & \underline{84.26}      & \underline{5.59}        \\
      TreeEnergy\cite{liang2022tree}     & 77.90$\pm$22.56                  & 14.51$\pm$16.25        & 71.69$\pm$6.38                 & 90.27$\pm$48.94     & 82.06$\pm$24.39 & 12.42$\pm$24.39  & 80.11$\pm$13.55 & 22.33$\pm$33.15  & 87.45$\pm$16.77 & 16.01$\pm$30.88 & 83.20      & 16.92       \\
      $\text{S}^2$ME \cite{wang2023s}   & 75.72$\pm$25.09                  & 14.78$\pm$19.45        & 73.14$\pm$5.06                 & 91.45$\pm$46.97     & \underline{83.10$\pm$24.17} & 6.09$\pm$8.95    & 76.51$\pm$15.96 & 53.78$\pm$56.26    & 87.83$\pm$16.66 & \textbf{3.69$\pm$6.97}   & 82.48      & 21.19        \\
      ScribbleVC \cite{li2023scribblevc} & \underline{78.33$\pm$25.63}                  & \textbf{13.68$\pm$19.09}        & 73.54$\pm$5.95                 & 71.72$\pm$41.85     & 82.84$\pm$25.35 & \textbf{5.49$\pm$7.66}    & 80.28$\pm$12.77 & 6.35$\pm$12.51   & 88.56$\pm$16.37 & 5.65$\pm$14.21  & 83.89      & 5.83        \\
      ProCNS (Ours)                                        & \textbf{79.49$\pm$23.33}$^{*}$                  & 14.34$\pm$19.96        & \textbf{76.11$\pm$4.86}$^{*}$                 & \textbf{69.11$\pm$40.15}$^{*}$     & \textbf{83.83$\pm$24.78}$^{*}$ & 7.52$\pm$15.27   & \underline{81.66$\pm$14.30} & \underline{5.51$\pm$10.78}   & \textbf{88.70$\pm$17.05} & 3.97$\pm$9.15   & \textbf{84.73}      & 5.66        \\ \hline
      $\text{UNet}_{F}$                                    & 82.25$\pm$20.13                  & 15.34$\pm$23.35        & 79.30$\pm$5.57                 & 63.86$\pm$39.74     & 84.88$\pm$25.30 & 4.51$\pm$7.65   & 83.76$\pm$15.22 & 4.07$\pm$8.27    & 89.40$\pm$18.25 & 3.81$\pm$8.74   & 86.01      & 4.13        \\ \hline
      \end{tabular}
  }
  \vspace{-0.1cm}
\end{table*}
\begin{figure*}[!h]
  \centering
  \includegraphics[width=1\textwidth]{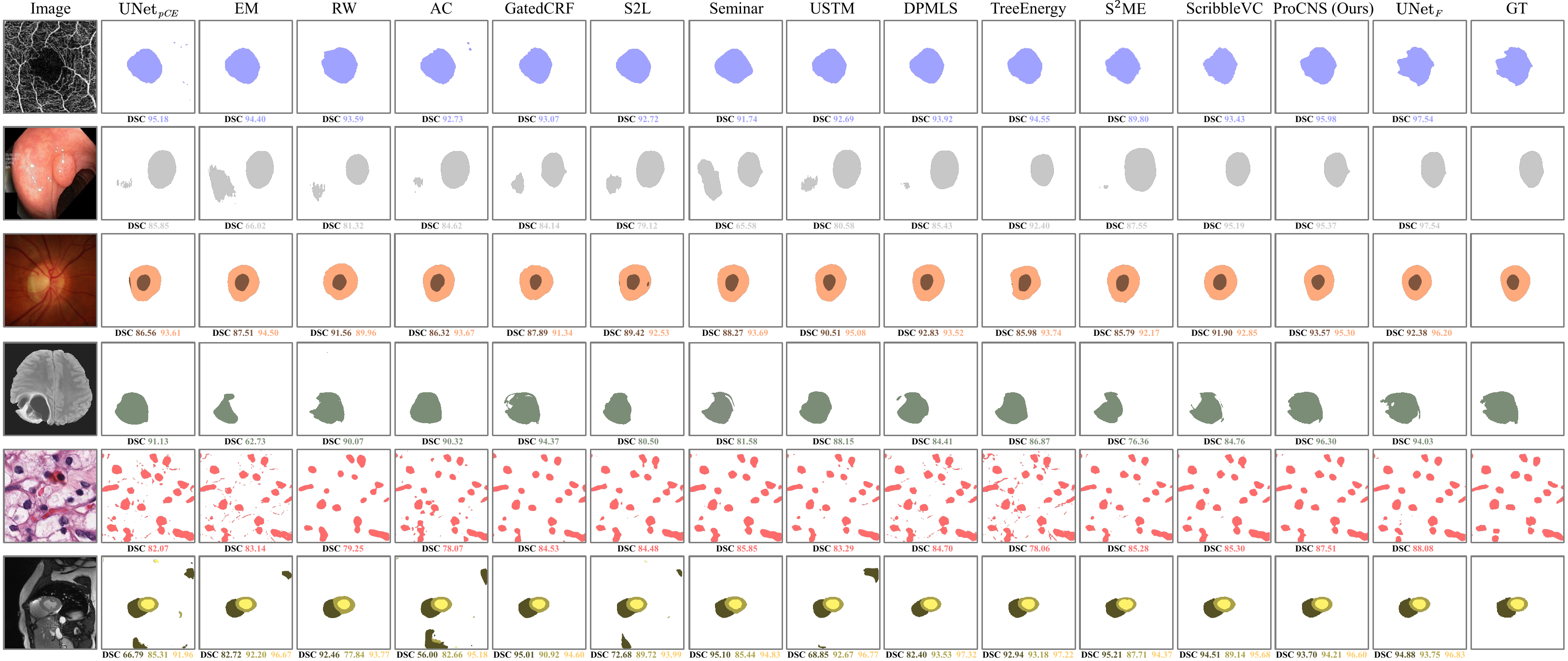} 
  \caption{Visualization of representative segmentation results from ProCNS and other SOTA WSS methods on the FAZ, Ployp, ODOC, WT, Nuclei and CM tasks.}
  \label{fig3_output_6}
\end{figure*}
\subsection{Comparison with State-of-the-arts}
To validate the superiority of the proposed algorithm, we compare ProCNS with a series of state-of-the-art WSS algorithms across the six distinct segmentation tasks. Among these compared methods, USTM \cite{liu2022weakly} exploits consistency learning techniques to regularize the training process. Seminar Learning (SeL) \cite{chen2021seminar} distills fine-grained semantic infromation from labeled pixels to guide the training of the final model. S2L \cite{lee2020scribble2label} and DMPLS \cite{luo2022scribble} propagate semantic information from labeled pixels to unlabeled ones to generate pseudo-labels. GatedCRF \cite{obukhov2019gated} and TreeEnergy \cite{liang2022tree} employ low-dimensional features from the original images for additional supervision.

In terms of computational burden, in Table \ref{tab:comparsion2_time} we present the training time and testing efficiency for various compared methods on the FAZ task. It is important to note that, for a fairness purpose, these metrics are provided under identical hardware and communication conditions. The results indicate that ProCNS has a slightly increased training time compared to certain WSS methods while maintaining comparable testing efficiency.

In terms of model performance, as shown in Table \ref{tab:comparison} and Table \ref{tab:comparison_add}, ProCNS outperforms almost all other compared methods on all the six segmentation tasks. The superiority of ProCNS could be due to the collaborative synergy between the PRSA loss and ANPM. Specifically, the PRSA loss leverages the calibrated prototypes updated by ANPM's noise masking mechanism to provide more precise guidance. On the other hand, ANPM utilizes the prediction refinement strategy of the PRSA loss to better perceive noisy regions. It is observed that the compared methods may be more suitable for segmenting relatively regular anatomical structures. When dealing with irregular pathological regions of interest, such as polyps, these methods exhibit subpar performance. In contrast, our approach exhibits superior generalization, not only excelling on the segmentation of the former but also delivering highly satisfactory performance on the latter. Specifically, for polyp segmentation, our proposed method significantly outperforms the second-best approach (EM), achieving a DSC improvement of 3.25\% at $\rm p \leq 0.05$. On FAZ segmentation, compared over the second-best method (TreeEnergy), ProCNS achieves a noteworthy DSC improvement of 0.9\% at $\rm p \leq 0.05$, with the difference from the fully-supervised DSC being merely 1.4\%. Our method holds a 0.51\% lead over the second-ranked method (EM) on OD segmentation at $\rm p \leq 0.05$. The average DSC of ODOC from ProCNS is 0.23\% higher than that from the second-best method (ScribbleVC). For WT segmentation, our method holds a 1.16\% lead over the second-ranked method (ScribbleVC) at $\rm p \leq 0.05$. For nuclei segmentation, our method significantly outperforms the second-best approach (GatedCRF), achieving a DSC improvement of 0.77\% at $\rm p \leq 0.05$. On the CM segmentation task, our method holds a 0.73\% lead over the second-ranked method ($\text{S}^2$ME) for the RV structure at $\rm p \leq 0.05$. The average DSC of RV, MYO and LV from ProCNS is 0.47\% higher than that from the second-best method (DMPLS).

\subsection{Verification of Seamless Plugin Integration}
\begin{table*}[t]
  \caption{Validation experiments of ProCNS's potential as a seamless integration plugin on the FAZ, Ployp and ODOC tasks. “$*$” represents $\rm p \leq 0.05$ in a Wilcoxon signed-rank test comparing the corresponding evaluation metric of the WSS method with and without ProCNS.}
  \label{tab:validation_results}
  \centering
  \renewcommand\arraystretch{1.2}
  \tabcolsep=0.2cm{
  \begin{tabular}{l|cc|cc|cc}
  \hline
  \multirow{2}{*}{\textbf{Method}}                            & \multicolumn{2}{c|}{\textbf{FAZ}}    & \multicolumn{2}{c|}{\textbf{Polyp}}   & \multicolumn{2}{c}{\textbf{ODOC}}    \\ \cline{2-7} 
                                                      & \textbf{DSC}$\uparrow$          & \textbf{HD95}$\downarrow$        & \textbf{DSC}$\uparrow$          & \textbf{HD95}$\downarrow$         & \textbf{DSC}$\uparrow$          & \textbf{HD95}$\downarrow$         \\ \hline
  pCE \cite{tang2018normalized}    & 90.02        & 29.02       & 78.99        & 50.02        & 87.39        & 14.02        \\
  pCE w/ ProCNS                                      & 92.60(+2.58)$^{*}$ & 7.43(-21.59)$^{*}$ & 80.51(+1.52)$^{*}$ & 48.89(-1.13)$^{*}$ & 87.62(+0.23) & 10.51(-3.51)$^{*}$ \\ \hline
  GatedCRF \cite{obukhov2019gated} & 92.39        & 7.49        & 79.24        & 46.35        & 87.72        & 10.24        \\
  GatedCRF w/ ProCNS                                 & 93.30(+0.91)$^{*}$ & 7.02(-0.47)$^{*}$ & 80.62(+1.38)$^{*}$ & 49.73(+3.38)$^{*}$ & 88.17(+0.45)$^{*}$ & 10.17(-0.07) \\ \hline
  TreeEnergy \cite{liang2022tree}  & 92.84        & 7.66        & 76.94        & 52.01        & 87.63        & 11.06        \\
  TreeEnergy w/ ProCNS                               & 93.22(+0.38) & 7.36(-0.30) & 82.37(+5.43)$^{*}$ & 51.45(-0.56)$^{*}$ & 87.93(+0.30) & 10.19(-0.87)$^{*}$ \\ \hline
  $\text{S}^2$ME \cite{wang2023s} & 92.33        & 12.44       & 77.26        & 57.91        & 87.82        & 10.24        \\
  $\text{S}^2$ME w/ ProCNS                          & 92.93(+0.60)$^{*}$ & 6.89(-5.55)$^{*}$ & 81.41(+4.15)$^{*}$ & 54.01(-3.90)$^{*}$ & 88.33(+0.51)$^{*}$ & 9.94(-0.30)  \\ \hline
  $\text{UNet}_{F}$                                                   & 95.14        & 4.80        & 86.01        & 48.72        & 88.54        & 9.61         \\ \hline
  \end{tabular}
  }
  \vspace{-0.4cm}
  \end{table*}
  \begin{figure}[!b]
    \vspace{-0.2cm}
        \centering
        \includegraphics[width=0.8\columnwidth]{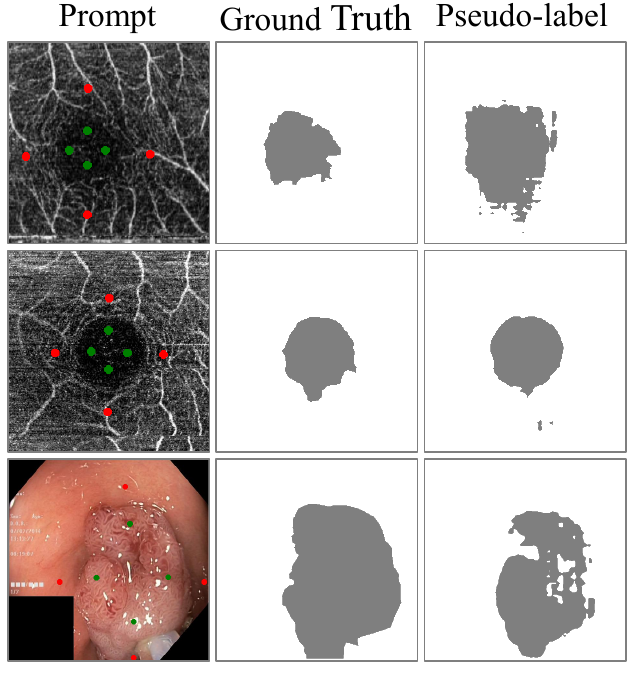}
        \caption{\hspace{-0.1cm}Visualization of pseudo-labels generated by SAM-Med2D \cite{cheng2023sam}. The overlaid green and red points respectively denote target and background prompts.}
        \label{fig:visual_samed2d}
    \vspace{-0.3cm}
\end{figure}

In this subsection, we aim to experimentally validate our statement that ProCNS's Main stage can serve as a seamlessly integrable plugin for other WSS methods, further enhancing model performance. Specifically, we replace the Initialization stage of ProCNS with four different WSS methods (including their models, losses and training paradigms) and train them until convergence is achieved (obtaining preliminary segmentation models). Subsequently, we proceed with the Main stage. As tabulated in Table \ref{tab:validation_results}, it is evident that compared to the original methods, the approaches with ProCNS as a follow-up plugin consistently demonstrate varying degrees of performance improvements across the three tasks. Notably, the integration of ProCNS with pCE \cite{tang2018normalized} results in a significant 3.68\% DSC improvement on the FAZ task compared to the original method. Similarly, the combination of either TreeEnergy \cite{liang2022tree} or $\text{S}^2$ME \cite{wang2023s} with ProCNS respectively yields DSC improvements of 5.43\% and 2.57\% over the original counterparts. This confirms the capability of ProCNS to act as a follow-up plugin, aiding WSS methods by mitigating the impact of noise accumulation and overconfident prediction, thereby enhancing the final segmentation performance.

\begin{table}[t]
  \vspace{-0.1cm}
\caption{Performance comparisons (DSC) with and without utilizing pseudo-labels generated by SAM-Med2D on the FAZ and polyp tasks.}
\centering
\label{tab:ablation_add_foundation}
\renewcommand\arraystretch{1.5}
  \begin{tabular}{c|ccc}
    \hline
    Method                       & FAZ    & Polyp \\ \hline

    SAM-Med2D w/ ProCNS                       & 93.30  & 79.04 \\

    ProCNS                                    & 93.74  & 82.21 \\ \hline
    \end{tabular}
\vspace{-0.2cm}
\end{table}
\subsection{Exploratory Analysis of Incorporating Foundation Models}
Benefiting from the development of SAM \cite{kirillov2023segment}, several studies \cite{yang2024foundation} have emerged that utilize foundation models to assist WSS. We assess the impact of integrating foundation models into ProCNS. Specifically, we exclude the Initialization stage and utilize predictions from a foundation model as the initial pseudo-labels in the Main stage. Recent research \cite{cheng2023sam, chen2023aslseg, ma2024segment} indicates that directly applying the pretrained SAM to medical image segmentation is suboptimal due to the significant domain gap between natural and medical images. Therefore, we employ SAM's medical variant, e.g., SAM-Med2D \cite{cheng2023sam}, to generate pseudo-labels for the FAZ and polyp tasks (as depicted in Fig. \ref{fig:visual_samed2d}). As shown in Table \ref{tab:comparison} and Table \ref{tab:ablation_add_foundation}, for those two FAZ and polyp tasks, utilizing pseudo-labels generated by SAM-Med2D (SAM-Med2D w/ ProCNS) yields better performance than most WSS methods. Despite its performance still lagging behind the original design of ProCNS, the results still indicate that leveraging foundation models to assist WSS is a promising direction.

\subsection{Analysis on the Impact of Annotation Sparsity}
\begin{table}[t]
  \vspace{-0.2cm}
  \caption{The performance of ProCNS with various rates of sparse and full annotations on the FAZ, Ployp and ODOC tasks. The best results are in bold.}
  \label{tab:data_sensitivity}
  \centering
  \renewcommand\arraystretch{1.2}
  \resizebox{\columnwidth}{!}{
  \begin{tabular}{cc|c|cc|cc|cc}
  \hline
  \multicolumn{2}{c|}{\multirow{2}{*}{\textbf{Method}}} & \multirow{2}{*}{\textbf{Sparse : Full}} & \multicolumn{2}{c|}{\textbf{FAZ}} & \multicolumn{2}{c|}{\textbf{ODOC}} & \multicolumn{2}{c}{\textbf{Polyp}} \\ \cline{4-9} 
                          &  &                      & \textbf{DSC}$\uparrow$        & \textbf{HD95}$\downarrow$        & \textbf{DSC}$\uparrow$        & \textbf{HD95}$\downarrow$        & \textbf{DSC}$\uparrow$         & \textbf{HD95}$\downarrow$        \\ \hline
  \multirow{5}{*}{Ours} & \#1 & 100\% : 0\%     & 93.74                           & 6.72 & 88.32                           & 9.83         & 82.73                           & 47.80                   \\ \cline{2-9} 
                          & \#2 & 67\% : 33\%  & \textbf{95.72}                & \textbf{4.58}  & 88.62                & 9.46     & 82.80                           & 50.91                \\ \cline{2-9} 
                          & \#3 & 50\% : 50\%  & 95.57                           & 4.71  & 88.92                           & 8.96    & 85.40                           & 47.36                       \\ \cline{2-9} 
                          & \#4 & 33\% : 67\%  & 95.30                           & 4.77  & 89.01                           & 8.73      & 85.97                           & 52.33                     \\ \cline{2-9} 
                          & \#5 & 0\% : 100\%  & 95.38                           & 4.79    & \textbf{89.74}                           & \textbf{8.36}      & \textbf{86.64}                           & \textbf{45.86}                     \\ \hline
  \multicolumn{2}{c|}{$\text{UNet}_{F}$}     & 0\% : 100\%  & 95.14            & 5.46    & 88.54                           & 9.61    & 86.01                           & 48.72                       \\ \hline
  \end{tabular}
  }
  \vspace{-0.2cm}
\end{table}

To explore the impact of the annotation sparsity level on ProCNS and inspired by Zhang \emph{et al.} \cite{zhang2022cyclemix}, we report the results of ProCNS obtained from training samples with five different sparse-versus-full annotated proportions. As shown in Table \ref{tab:data_sensitivity}, ProCNS demonstrates superior performance across all five proportions, all of which surpass the performance of $\text{UNet}_{F}$. For the ODOC and Polyp tasks, as expected, the performance of ProCNS increases as the proportion of full annotations increases. The segmentation performance respectively reaches its peak DSC of 89.74\% and 86.64\% at the 100\% full annotation. For the FAZ task, it is notable that at approximately a 30\% full annotation, ProCNS reaches a peak DSC of 95.72\%. Further increasing the proportion leads to slight declines in model performance. This suggests that higher proportions of full annotations do not necessarily lead to better performance. Such phenomenon may be attributed to the high sensitivity of sample-wise prototypes within ProCNS and the inherent subjectivity (or noise) presented in full annotations, especially for regions of interest with ambiguous boundaries. Therefore, as the proportion increases, some prototypes might be affected by noise and deviate from accuracy, leading to a decline in model performance. Nonetheless, these results still indicate that providing more annotations to ProCNS can lead to additional performance improvements.
\begin{figure}[b]
  \centering
  \footnotesize
  \vspace{-0.4cm}
    \subfigure{\includegraphics[width=0.33\linewidth]{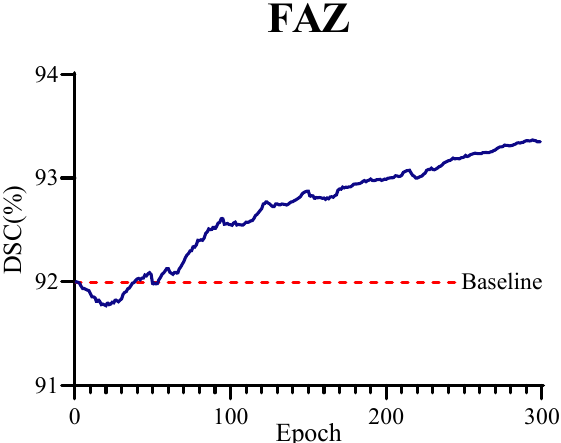}}\hspace{-0.08cm}
      \subfigure{\includegraphics[width=0.33\linewidth]{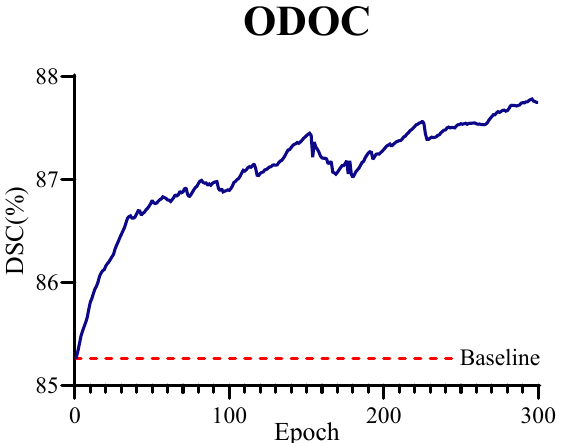}}\hspace{-0.08cm}
      \subfigure{\includegraphics[width=0.33\linewidth]{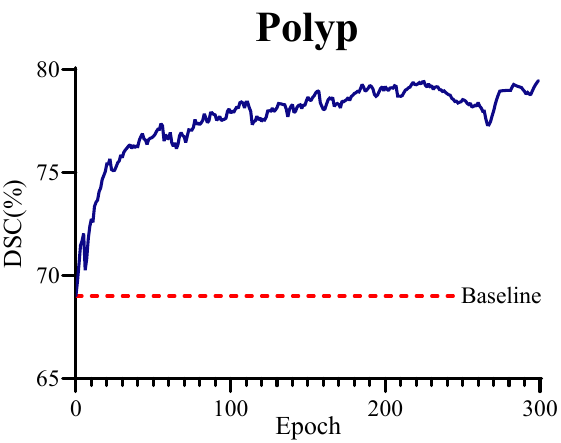}}\hspace{-0.08cm}
      \\
  \caption{Quantitative analysis on noise suppression concerning the DSC between the denoised labels and corresponding full ground truth. The baseline denotes the DSC between their initial pseudo-labels and corresponding full ground truth.}
  \label{fig:noise_sup}
\end{figure} 
\begin{figure}[t]
  \vspace{-0.2cm}
      \centering
      \includegraphics[width=1\columnwidth]{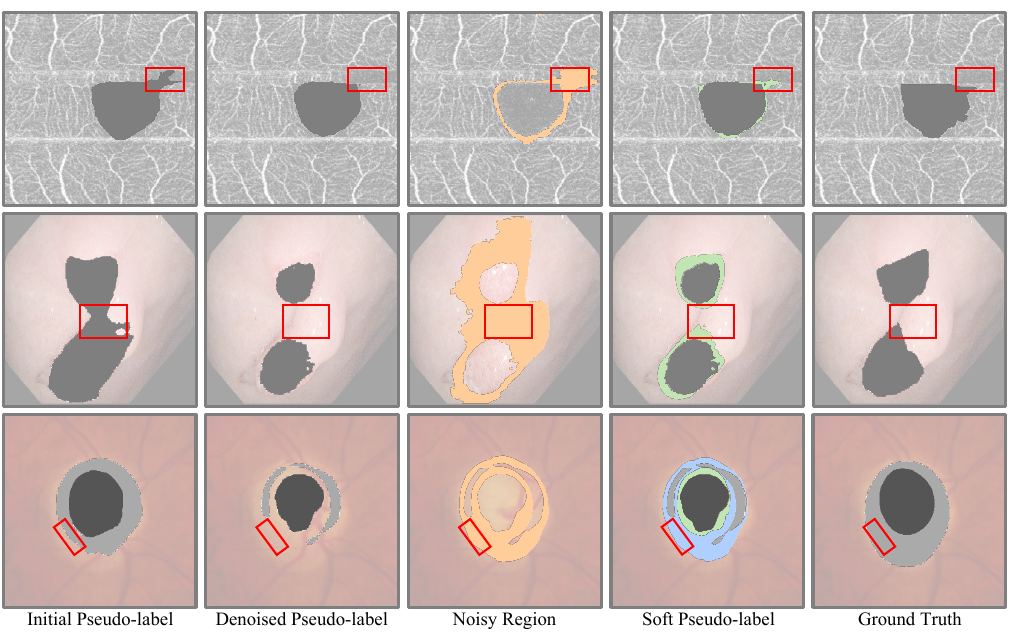}
      \caption{Visualization of noise suppression. The overlaid orange denotes masked noisy regions. The overlaid green and blue denote additional soft pseudo-labels generated through reassignment. The regions where obvious labeling errors are evident in the initial pseudo-labels are highlighted with red boxes.}
      \label{fig:visual_suppression}
  \vspace{-0.4cm}
  \end{figure}
\subsection{Effectiveness Analysis on Noise Suppression}
The ablation studies in \ref{subsec:ab_procns} show that ANPM plays a crucial role in achieving outstanding performance. This is credited to ANPM's adeptness in iteratively identifying and masking noisy regions within the pseudo-labels. In this subsection, we perform the effectiveness analysis on the noise suppression efficacy of ANPM. Concretely, we first evaluate the DSC between denoised labels and their corresponding full ground truth at various training epochs. As illustrated in Fig. \ref{fig:noise_sup}, with an increase in epochs, there are consistent increases in DSC across all three tasks. This affirms ANPM's effectiveness in accurately identifying and suppressing noisy regions. To further validate the role of ANPM, we visualize the masked noisy regions alongside their respective soft pseudo-labels for the three tasks. As shown in Fig. \ref{fig:visual_suppression}, for the FAZ and polyp tasks, noisy regions predominantly reside at the boundaries of the segmentation targets. As for the ODOC task, a substantial amount of noise is presented at the OD and OC intersection boundary. The soft pseudo-labels generated for these regions through the reassignment strategy are of high quality (as depicted in the red boxes of Fig. \ref{fig:visual_suppression}). This reaffirms the significant role of ANPM. It also elucidates the motivation behind utilizing these unique pseudo-labels for additional supervision over noisy regions.

\section{Discussion}

This study presents and evaluates a novel WSS algorithm, named ProCNS. ProCNS diverges from recent WSS algorithms \cite{wu2023compete, liang2022tree, wu2024reliability, zhang2022cyclemix, wang2023s} that predominantly rely on global context (e.g., position, topology) and predefined segmentation cues (e.g., intensity) from sparse annotations. Instead, it concentrates on the potentially overlooked characteristic of sparse annotations generated in practice; namely, the fact that such annotations tend to occur in less-informative regions rather than hard-yet-informative ones. Its core design principle lies in utilizing sparse annotations alongside the immense potential of segmentation networks to adaptively discern between less-informative regions and hard-yet-informative ones, subsequently providing more fine-grained and specialized supervision for hard-yet-informative regions. Specifically, we integrate the concepts of prototype-based nearest-neighbor searching and pair-wise affinity to formulate $\boldsymbol{\mathcal{L}}_{\rm \textbf{PRSA}}$, providing the model with accurate semantic guidance. Then, taking into account the fact that prototypes obtained directly from sparse annotations lack semantic richness and accuracy, we employ prototype calibration to design the ANPM module. Additionally, based on the strategies of reassignment and soft supervision, we devise $\boldsymbol{\mathcal{L}}_{\rm \textbf{noise}}$ for the noisy regions pinpointed by ANPM, offering supplementary supervision.

\subsection{Synergistic Interplay of Components}
As shown in Table \ref{tab:comparison} and Fig. \ref{fig3_output_6}, our ProCNS achieves superior performance. We attribute this success to the synergistic interplay of $\boldsymbol{\mathcal{L}}_{\rm \textbf{PRSA}}$, ANPM and $\boldsymbol{\mathcal{L}}_{\rm \textbf{noise}}$. Specifically, $\boldsymbol{\mathcal{L}}_{\rm \textbf{PRSA}}$ utilizes ANPM's noise masking mechanism to calibrate prototypes, providing more accurate semantic guidance. In return, ANPM benefits from the refinement strategy of $\boldsymbol{\mathcal{L}}_{\rm \textbf{PRSA}}$, better perceiving noisy regions. On top of this foundation, $\boldsymbol{\mathcal{L}}_{\rm \textbf{noise}}$ is crafted to provide extra soft supervision for the noisy regions. Such effect is clearly observed in Table \ref{tab:ablation1} and Fig. \ref{fig:ablation_refinement}.

\subsection{Prototype Calibration}
Prototype learning has been widely investigated in segmentation tasks. Conventional prototype-based segmentation algorithms \cite{cheng2024few, ding2023few, sun2022few} primarily utilize a MAP operation to generate support prototypes for a specific class of interest, which are subsequently employed to predict the segmentation masks. Recently, within the domain of few-shot segmentation, several studies have innovatively explored prototype calibration and refinement, significantly enhancing model performance. Zhu \emph{et al.} \cite{zhu2023few} design a Region-enhanced Prototypical Transformer (RPT) to generate multiple prototypes for various regions within the target, integrating them to obtain an ideal prototype; Cheng \emph{et al.} \cite{cheng2024few} generate extra boundary prototypes for foreground and background by applying automated processes, such as erosion and dilation, on the ground truth. Nevertheless, the aforementioned approaches come with their drawbacks: the former inevitably increases the computational burden; the latter introduces extra hyperparameters via the automated algorithms, which makes it more time-consuming to optimize and tune the model. In contrast, our prototype calibration strategy leverages the model's potential to progressively detect noisy regions within the mask (the initial pseudo-labels) and then employs MAP to secure accurate prototypes, without notably increasing the computational burden or the model's complexity. The effectiveness has been established in Fig. \ref{fig:noise_sup} and Fig. \ref{fig:visual_suppression}.

We expand our exploration of prototype granularity and scale in Table \ref{tab:ablation3_prototype}. To the best of our knowledge, this is the first attempt within the realm of WSS to utilize progressive prototype calibration and noise suppression to tackle the deficiency of prototype semantic representativeness and richness.

\subsection{Impacts and Future Works}

This work holds promise of facilitating many relevant studies aiming at label-efficient through pseudo-proposal and prototype calibration, encompassing but not limited to weakly supervised learning, semi-supervised learning, noise learning and few-shot learning. Notably, as depicted in Table \ref{tab:validation_results}, ProCNS can serve as a seamless integration plugin to help other WSS methods break through performance bottlenecks. Furthermore, as shown in Table \ref{tab:data_sensitivity}, a higher proportion of full annotations does not necessarily guarantee better performance, which seems counter-intuitive. We conjecture this might be related to the quantity of the training data and the inevitable subjective noise in the ambiguous boundaries of the manual full annotations. Nonetheless, the result that ProCNS outperforms the model solely employing the CE loss when trained under 100\% full annotation demonstrates that ProCNS is capable of mitigating the aforementioned issue to a certain degree. However, effectively and efficiently classifying the most hard-yet-informative regions with inexact annotations remains an open question in our study. Foundational segmentation models pretrained on a large-scale dataset, e.g., SAM \cite{kirillov2023segment} and SAM-Med2D \cite{cheng2023sam}, demonstrate impressive zero-shot segmentation capabilities and superior performance without requiring additional training, which holds promise in addressing the aforementioned issues (as delineated in Table \ref{tab:ablation_add_foundation}). Looking forward, reasonably leveraging predictions from foundation models to select high-informative regions for fine-grained supervision represents a promising research direction.
\begin{figure}[t]
      \centering
      \includegraphics[width=1\columnwidth]{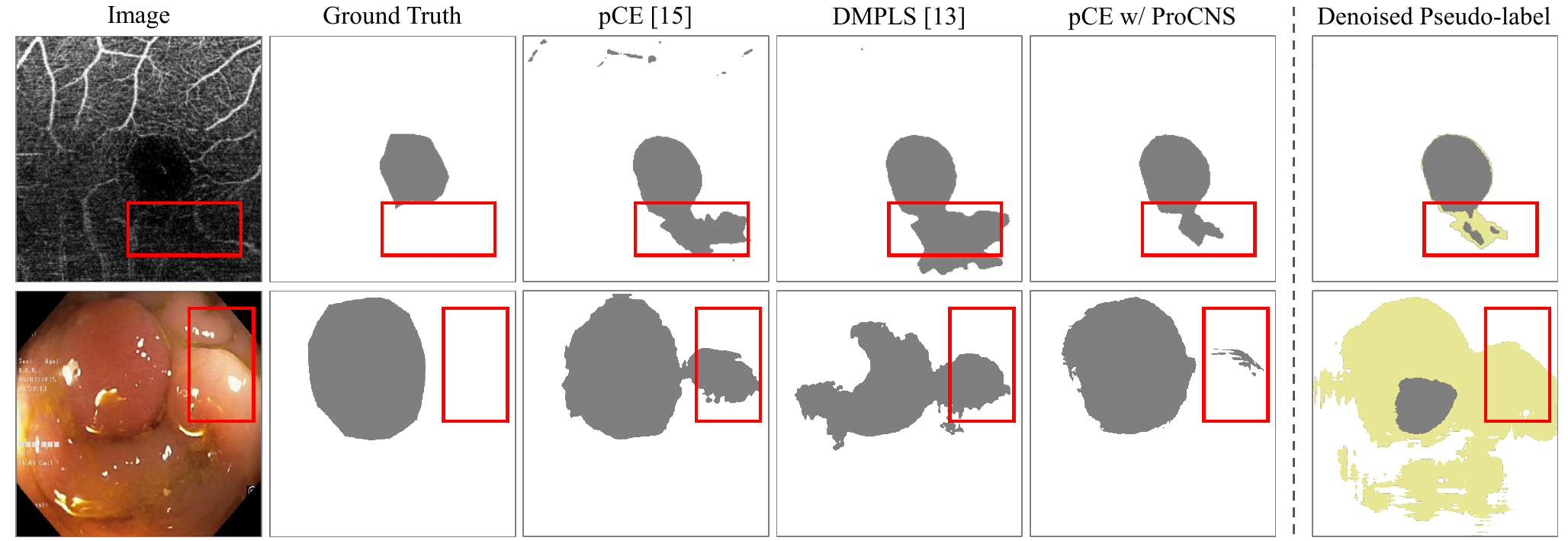}
      \caption{Visualization of several failure cases. The overlaid orange denotes masked noisy regions. The obvious prediction errors are highlighted with red boxes.}
      \label{fig:visual_limitation}
\end{figure}
\subsection{Limitations}
Concerning our method's limitation, ProCNS discerns hard-yet-informative regions based on the degree of prediction volatility for ambiguous regions by the model itself. This means it heavily relies on the model's potential to identify such regions. If the model lacks sufficient discriminative strength for a particular task, the identified informative regions may be meaningless and may even exacerbate the model's bias. To better illustrate the aforementioned limitation, we train ProCNS employing only the pCE loss at the Initialization stage to diminish the discriminative strength of the preliminary model and then visualize some failure cases. As shown in Fig. \ref{fig:visual_limitation}, ProCNS, like other WSS methods, makes false positive errors for some difficult-to-segment images. The first case demonstrates that when the preliminary model overfits certain non-target regions, the ANPM module identifies a reduced number of noisy regions. The denoised pseudo-labels contain over-confident errors, leading to cumulative model errors. The second case indicates that when the preliminary model exhibits a high degree of prediction volatility, the ANPM module identifies noisy regions much larger than the reliable regions. Consequently, the denoised pseudo-labels are severely eroded, significantly diminishing the semantic richness and accuracy of the calibrated prototypes. This leads to performance degradation. In these two cases, enhancing model's fundamental discriminability might be necessary, for instance, by employing more advanced network architectures or more appropriate loss functions. The aforementioned conjecture is further evidenced by the fact, as shown in Table \ref{tab:ablation_add_initialprsaloss}, that all results without the initial PRSA loss exhibit significant performance gaps compared to those with it, particularly in the relatively challenging polyp lesion segmentation task.

Concerning our experiments' limitation, given the constraints on the computational resources and the inclusion of multiple tasks and scenarios, all ProCNS-related experiments employ the vanilla UNet rather than more advanced networks. This decision is supported by previous studies \cite{lin2022yolocurvseg, isensee2021nnu} indicating that in the field of medical image segmentation, the vanilla UNet does not exhibit significant performance gaps compared to other advanced networks.

\section{Conclusion}
This paper proposes a novel WSS framework for medical image segmentation, named ProCNS, which can effectively alleviate model degradation caused by representation bias and noise accumulation. Extensive experiments are conducted on FAZ, ODOC, polyp, nuclei, cardiac multi-structures and whole brain tumor segmentation tasks, with the superiority of our proposed framework being successfully established.
    \bibliography{ProCNS}
    \bibliographystyle{IEEEtran}

\end{sloppypar}
\end{document}